\documentclass{article}

\makeatletter
\providecommand{\@trackname}{NeurIPS Conference.}
\makeatother

\usepackage[final]{neurips_2026} 
\usepackage[utf8]{inputenc}
\usepackage[T1]{fontenc}
\usepackage[hidelinks,colorlinks=false,allcolors=black,pdfborder={0 0 0}]{hyperref}
\usepackage{url}
\usepackage{booktabs}
\usepackage{amsfonts}
\usepackage{nicefrac}
\usepackage{microtype}
\usepackage[dvipsnames]{xcolor}

\usepackage[dvipsnames]{xcolor}                              
\usepackage{amsmath,amssymb,amsthm}
\usepackage{algorithm}
\usepackage{algpseudocode}
\usepackage{booktabs}
\usepackage{graphicx}
\usepackage{hyperref}
\usepackage{natbib}
\setcitestyle{numbers,square,sort\&compress}
\usepackage{tikz}
\usetikzlibrary{positioning,arrows.meta,fit,backgrounds,calc}

\title{LAPO: KL-Free Logit-Averaged Policy Optimization to Leverage RL and SFT in LLM Post-Training\thanks{Accepted to the NeurIPS 2026 Workshop on \emph{Transitioning from Pre-Training to Post-Training}. First released on arXiv on May 19, 2026. Code is publicly available at \url{https://github.com/xingweiGan/LAPO}. The authors are grateful to the anonymous reviewers for their constructive comments.}}

\author{%
 Xingwei Gan\quad{} Ying Zhu\thanks{Ying Zhu thanks UCSD senate research fund and annual research fund for supporting this work.} \\
  University of California San Diego \\
  \texttt{xigan@ucsd.edu\quad{}yiz012@ucsd.edu} \\
}

\date{}

\begin{document}
\maketitle

\begin{abstract}
From a classical statistical perspective, model averaging and parameter shrinkage (e.g., Ridge regression with an $l_2$-penalty) are deeply connected mechanisms designed to reduce estimator variance and lower mean squared error. Motivated by this insight, we introduce Logit-Averaged Policy Optimization (LAPO), a KL-free and critic-free post-training framework that replaces explicit KL regularization with logit averaging inside Group Relative Policy Optimization (GRPO); in particular, averaging the logits of the trainable policy and a frozen reference policy functions as an implicit structural regularizer that stabilizes rollout trajectories. For selecting the mixing weight, we propose a compute-efficient adaptive algorithm that dynamically balances exploration and structural anchoring throughout training, evaluated against a static fixed-weight baseline. This adaptive mechanism eliminates explicit KL penalty backpropagation and expensive, fragile hyperparameter sweeps while reducing the incidence of uninformative zero-reward rollout groups. Mathematically, embedding logit averaging within the reinforcement learning loop operates as an in-training Product of Experts (PoE): the multiplicative geometric mean enables the reference anchor to act as a soft logical ``AND'' that vetoes non-compliant tokens, enforcing schema rigor and safety guardrails without the catastrophic forgetting typical of sequential fine-tuning. Evaluated across Qwen2.5-Instruct models (1.5B, 3B, and 7B), LAPO consistently outperforms canonical KL-regularized GRPO, achieving a 7\%--15\% accuracy gain on MATH and a 2\%--5\% gain on cn\_k12 and MMLU. Finally, the resulting mixed teacher policy can be distilled back into a single student network, fully restoring standard $1\times$ inference latency.

\end{abstract}



\section{Introduction}

Post-training large language models (LLMs) fundamentally relies on two paradigms: supervised fine-tuning (SFT) on curated demonstrations and reinforcement learning (RL) from human feedback (RLHF; \cite{christiano2017preferences,ouyang2022instructgpt,bai2022rlhfanthropic}) or verifiable rewards (RLVR; \cite{grpo,guo2025deepseekr1}). While SFT provides dense, token-level supervision, it is inherently bounded by the distribution of its training demonstrations. Conversely, RL algorithms such as PPO~\citep{ppo}, GRPO~\citep{grpo}, and DAPO~\citep{dapo} overcome these demonstration boundaries by exploring reward landscapes, albeit under sparse and noisy feedback. The standard post-training pipeline chains these paradigms sequentially: SFT first establishes formatting and domain competence, followed by RL with a Kullback-Leibler (KL) penalty 
to keep the trainable policy $\pi_\theta$ close to the frozen SFT reference policy $\pi_{\mathrm{ref}}$.

However, this KL regularization paradigm introduces a fundamental tension: pulling $\pi_\theta$ too close to $\pi_{\mathrm{ref}}$ forces the trainable policy to inherit the reference model's weaknesses---such as dropped algebraic substitutions, repetition loops and narrow solution coverage---while drifting too far invites reward hacking and format degradation. Preserving safety priors and structural schema without overly restricting policy exploration thus requires delicate, compute-intensive sweeps over the penalty hyperparameter. 

To resolve this bottleneck, we introduce \textbf{Logit-Averaged Policy Optimization (LAPO)}, a KL-free and critic-free post-training framework that re-architects reference anchoring. Instead of penalizing policy divergence during optimization, LAPO directly averages the logits of the trainable policy $\pi_\theta$ and a frozen reference anchor $\pi_{\mathrm{ref}}$ using an adaptive schedule that dynamically adjusts the mixing weight during training (with a static fixed-weight schedule serving as the baseline).

We evaluate LAPO across three benchmarks---MATH~\citep{hendrycks2021math}, cn-k12, and MMLU~\citep{hendrycks2021mmlu}---and three model scales: Qwen2.5-Instruct (1.5B, 3B, and 7B). Across all nine $(\text{dataset} \times \text{model-size})$ evaluation cells, LAPO consistently outperforms canonical KL-regularized GRPO, achieving a 7\%--15\% accuracy gain on MATH and a 2\%--5\% gain on cn-k12 and MMLU.

\section{KL-Free Logit-Averaged Policy Optimization}

\label{sec:method}
Our proposal averages the logits of a reference policy (e.g., 
$\pi_{\mathrm{sft}}$) and a 
trainable policy $\pi_\theta$ parameterized by $\theta$. Given a weight
$\alpha\in[0,1]$ and state $s$ (i.e., a question or a ``partial" response), we define the \emph{mixed} logit by
\begin{align}
\label{eq:logit-mix}
z^{\mathrm{mix}}_{\theta,\alpha}(s) \;=\; (1-\alpha)\,z_\theta(s) + \alpha\,z_{\mathrm{ref}}(s),
\end{align} where $z_\theta(s)$ and $z_{\mathrm{ref}}(s)$ correspond to the logit vectors of the 
trainable policy and the reference policy at $s$. Then, the language model of our interest takes the form
\begin{equation}
\label{eq:mix policy}
\pi^{\mathrm{mix}}_{\theta,\alpha}(a\mid s) =\mathrm{softmax}\bigl(z^{\textrm{mix}}_{\theta,\alpha}(s)\bigr)(a)
= \frac{\exp z^\textrm{mix}_{\theta,\alpha}(s)_a}{\sum_{a'\in\mathcal{V}}\exp z^\textrm{mix}_{\theta,\alpha}(s)_{a'}},
\end{equation}
where $z^\textrm{mix}_{\theta,\alpha}(s)_{a'}$ denotes the value at $a'$ of the vector $z^\textrm{mix}_{\theta,\alpha}(s)$. Letting $x$ be a question and $y=(y_1,\dots,y_{|y|})$ be a response, the full sequence can be computed auto-regressively as
\begin{equation}
\pi^{\textrm{mix}}_{\theta,\alpha}(y\mid x) = \prod_{t=1}^{|y|} \pi_{\theta,\alpha}^{\textrm{mix}}(y_t\mid x, y_{<t}).
\end{equation}
We work with a verifiable reward $r:(x,y)\mapsto\{0,1\}$ that returns $1$ when $y$ provides the golden answer to $x$ 
in the correct format. Throughout the process, we only train $z_\theta$ (the logits of 
$\pi_\theta$) while freezing the logits $z_{\mathrm{ref}}$.
We generate the rollouts token-by-token by sampling 
$y_t\!\sim\!\pi_{\theta,\alpha}^{\mathrm{mix}}(\cdot\mid x, y_{<t})$.



\paragraph{Training objective.} 
We adopt GRPO \citep{grpo} and tailor it to the mixed policy. At the start of every
rollout collection, as in Algorithm~1 of~\cite{grpo}, line~6, we freeze a snapshot $\theta_{\mathrm{old}}$ of the trainable
parameters. We draw $G$ rollouts from the \emph{old} mixed policy,
$y_{i}\!\sim\!\pi^{\mathrm{mix}}_{\theta_{\mathrm{old}},\alpha}(\cdot\mid x)$, and form the group-relative advantage
\begin{align}
\label{eq:adv}
A_i \;=\; \frac{R_i - \overline{R}}{\mathrm{std}(R_1,\dots,R_G) + \epsilon},
\qquad
\overline{R} \;=\; \tfrac{1}{G}\sum_{j=1}^{G} R_j,
\end{align}
where $A_{i,t}\!=\!A_i$ for all
$t$, $R_i\!=\!r(x,y_i)\!\in\!\{0,1\}$, and $\epsilon$ is small enough for numerical stability. The per-token importance ratio between the mixed policies at the
current and old parameters takes the form
\begin{align}
\label{eq:ratio}
\rho_{i,t}(\theta,\alpha)
\;=\;
\frac{\pi^{\mathrm{mix}}_{\theta,\alpha}(y_{i,t}\mid x,y_{i,<t})}
     {\pi^{\mathrm{mix}}_{\theta_{\mathrm{old}},\alpha}(y_{i,t}\mid x,y_{i,<t})}.
\end{align}
With the clip radius $\varepsilon\!>\!0$, the training loss is defined as a
function of \emph{both} $\theta$ and $\alpha$:
\begin{equation}
\label{eq:loss}
\small
\mathcal{L}(\theta,\alpha,\theta_{\mathrm{old}})
= \textcolor{blue}{-}\,\mathbb{E}_{x\sim\mathcal{D}}\,
\mathbb{E}_{\{y_i\}\sim\pi_{\theta_{\mathrm{old}},\alpha}^{\mathrm{mix}}(\cdot\mid x)}
\left[
\frac{1}{\sum_i |y_i|}
\sum_{i=1}^{G}\sum_{t=1}^{|y_i|}
\min\!\Bigl(
\rho_{i,t}(\theta,\alpha)A_{i,t},
\mathrm{clip}\bigl(\rho_{i,t}(\theta,\alpha),1-\varepsilon,1+\varepsilon\bigr)A_{i,t}
\Bigr)
\right].
\end{equation}

Our framework requires neither KL regularization nor a critic network; the trainable policy and the reference
anchor are coupled through the logit averaging that defines
$\pi^{\mathrm{mix}}_{\theta,\alpha}$ with the weight $\alpha$ controlling the strength of the reference anchor. Moreover, our rollouts are generated from the \emph{mixed} policy $\pi^{\mathrm{mix}}_{\theta,\alpha}$ rather than from $\pi_\theta$; as a result, the reference anchor explicitly co-steers token generation during rollout, preventing format degradation during exploration.
Accordingly, both the numerator and the denominator of the importance ratio are evaluated at $\pi^{\mathrm{mix}}_{\theta,\alpha}$ and $\pi^{\mathrm{mix}}_{\theta_{\mathrm{old}},\alpha}$ rather than at $\pi_\theta$ and $\pi_{\theta_{\mathrm{old}}}$.

Conceptually, in KL regularization, $\pi_\theta$ gravitates toward $\pi_{\mathrm{ref}}$ and its limitations constrain the remaining gradient updates of $\theta$. On the other hand, our framework has the potential to let the future updates of $\theta$ 
overcome the limitations of $\pi_{\mathrm{ref}}$ while maintaining its formatting advantage through the logit averaging structure. We discuss this complementarity further in Section~\ref{sec:mechanism}. While our formal exposition focuses on GRPO due to its critic-free efficiency, replacing KL regularization with logit averaging is paradigm-agnostic.


In the following, we discuss two schemes for selecting the mixing weight: fixed-weight averaging (Algorithm~\ref{alg:fixed}) as a static baseline, and adaptive-weight averaging (Algorithm~\ref{alg:adaptive}).



\subsection{Fixed-mixing-weight baseline}
\label{sec:fixed}
A baseline is to fix $\alpha$ at a prescribed value
$\alpha_0\in(0,1)$ throughout the training process, where the objective is reduced to
$\mathcal{L}(\theta, \alpha_0, \theta_{\mathrm{old}})$.
See Algorithm~\ref{alg:fixed} for the detail. At each step, rollouts are sampled from $\pi^{\textrm{mix}}_{\theta_{\mathrm{old}},\alpha_0}(\cdot\mid x)$,
group-relative advantages are computed according to Eq.~\eqref{eq:adv}, and 
$\theta$ is updated through gradient steps with $\alpha\!=\!\alpha_0$ in Eq.~\eqref{eq:loss}.
Freezing the reference policy incurs a constant shift in the logit space.  

\begin{algorithm}[h]
\caption{Logit averaging with fixed mixing weight (baseline)}
\label{alg:fixed}
\begin{algorithmic}
\State \textbf{Input} frozen logits $z_{\mathrm{ref}}$ from a reference policy; pretrained base model logits $z_{\mathrm{base}}$; trainable logits $z_\theta \leftarrow z_{\mathrm{base}}$; fixed $\alpha_0 \in (0,1)$; prompt dataset $\mathcal{D}$; reward $r$; hyperparameters $T, \mu, G, \varepsilon, \eta$.
\For{$t = 1,\dots,T$}
  \State Sample a batch $\mathcal{D}_b$ from $\mathcal{D}$.
  \State Set $\theta_{\mathrm{old}} \leftarrow \theta$.
  \State Sample $G$ outputs $\{y_i\}_{i=1}^{G} \sim \pi^{\mathrm{mix}}_{\theta_{\mathrm{old}},\alpha_0}(\cdot \mid x)$ for each $x \in \mathcal{D}_b$.
  \State Compute rewards $\{R_i\}_{i=1}^{G}$ for each $y_i$ via the verifier reward $r$.
  \State Compute $A_{i,t}=A_i$ for the $t$-th token of $y_i$ through the group-relative advantage estimation (Eq.~\eqref{eq:adv}).
  \For{$m = 1,\dots,\mu$}
    \State Update $\theta$ by minimizing $\mathcal{L}(\theta,\alpha_0,\theta_{\mathrm{old}})$ (Eq.~\eqref{eq:loss}).
  \EndFor
\EndFor
\State \textbf{Output} $\pi_{\theta_{\mathrm{final}}}$ and $\pi^{\mathrm{mix}}_{\theta_{\mathrm{final}},\alpha_0}$.
\end{algorithmic}
\end{algorithm}

\subsection{Adaptive-mixing-weight: Compute-efficient dynamic tuning}
\label{sec:adaptive}
Our Algorithm ~\ref{alg:adaptive} updates $\alpha$ adaptively with the help of a
held-out validation set $\mathcal{D}_{\mathrm{val}} = \{(x_i, y^\star_i)\}_{i=1}^{N}$.
After each RL step, we decode two responses per validation prompt as follows,
\begin{align}
\label{eq:adapt-decode}
\hat y_i^{(\theta)} \;=\; \arg\max_y \pi_\theta(y\mid x_i),
\qquad
\hat y_i^{(\mathrm{mix})} \;=\; \arg\max_y \pi^{\mathrm{mix}}_{\theta,\alpha}(y\mid x_i).
\end{align}
Define the correctness indicators
$c_\theta(i) = r(x_i, \hat y_i^{(\theta)})$ and
$c_{\mathrm{mix}}(i) = r(x_i, \hat y_i^{(\mathrm{mix})})$, as well as
\begin{align}
\label{eq:adapt-counts}
s_1 \;=\; \#\bigl\{i:\, c_\theta(i){=}1,\ c_{\mathrm{mix}}(i){=}0\bigr\}
\;\;\text{(losing)},
\qquad
s_2 \;=\; \#\bigl\{i:\, c_\theta(i){=}0,\ c_{\mathrm{mix}}(i){=}1\bigr\}
\;\;\text{(gain)}.
\end{align}
We then introduce the net-gain metric $s_3 = (s_2-s_1-c_o)/c_d$ with offset $c_o$ and
scale $c_d$, and update the weight by
\begin{align}
\label{eq:adapt-update}
\alpha \;\leftarrow\; \sigma(s_3) \;=\; \frac{1}{1 + e^{-s_3}} \;\in\;(0,1).
\end{align}
When $s_2 \!>\! s_1 \!+\! c_o$ (i.e., mixing is net-helpful), the sigmoid function pushes
$\alpha$ closer to $1$; otherwise it pushes $\alpha$ closer to $0$.\footnote{In our experiments, we find $(c_o,c_d, |\mathcal{D}_{\mathrm{val}}|)=(25,35,100)$ gives the best performance.}

Each outer iteration $t$ performs $|\mathcal{D}_b| \cdot G$ rollout decodes (which are also required by Algorithm~\ref{alg:fixed} and the canonical GRPO) plus an additional $2 \cdot |\mathcal{D}_{\mathrm{val}}|$ greedy decodes for the adaptive step (one from $\pi_\theta$ and one from $\pi^{\mathrm{mix}}_{\theta,\alpha}$ on each validation prompt) to obtain $s_1$ and $s_2$. In our experiments, $|\mathcal{D}_b| \cdot G = 256 \times 8 = 2048$ and $|\mathcal{D}_{\mathrm{val}}| = 100$, so the validation step adds only $2 \cdot 100 / 2048 \approx 10\%$ overhead on top of the rollout decode cost. To further reduce this overhead in resource-constrained environments, future optimizations could be achieved by approximating the net-gain metric directly from in-batch training rollouts, freezing $\alpha$ early once it stabilizes, restricting validation to a minimal subset of boundary problems, or executing validation passes using lower precision (e.g., INT8) and capped sequence lengths.

\begin{algorithm}[h]
\caption{Logit averaging with adaptive mixing weights}
\label{alg:adaptive}
\begin{algorithmic}
\State \textbf{Input} frozen logits $z_{\mathrm{ref}}$ from a reference policy; pretrained base model logits $z_{\mathrm{base}}$; trainable logits $z_\theta \leftarrow z_{\mathrm{base}}$; prompt dataset $\mathcal{D}$; validation set $\mathcal{D}_{\mathrm{val}}$; reward $r$; offset $c_o$, scale $c_d$, initial weight $\alpha_0 \in (0,1)$; other hyperparameters as in Alg.~\ref{alg:fixed}.
\State $\alpha \leftarrow \alpha_0$.
\For{$t = 1,\dots,T$}
  \State Sample a batch $\mathcal{D}_b$ from $\mathcal{D}$.
  \State Set $\theta_{\mathrm{old}} \leftarrow \theta$.
  \State Sample $G$ outputs $\{y_i\}_{i=1}^{G} \sim \pi^{\mathrm{mix}}_{\theta_{\mathrm{old}},\alpha}(\cdot \mid x)$ for each $x \in \mathcal{D}_b$.
  \State Compute rewards $\{R_i\}_{i=1}^{G}$ for each $y_i$ via the verifier reward $r$.
  \State Compute advantages $A_{i,t}=A_i$ (Eq.~\eqref{eq:adv}).
  \For{$m = 1,\dots,\mu$}
    \State Update $\theta$ by minimizing $\mathcal{L}(\theta,\alpha,\theta_{\mathrm{old}})$ (Eq.~\eqref{eq:loss}).
  \EndFor
  \State Greedy-decode $\hat y_i^{(\theta)}$ and $\hat y_i^{(\mathrm{mix})}$ for each $x_i \in \mathcal{D}_{\mathrm{val}}$ (Eq.~\eqref{eq:adapt-decode}).
  \State $s_1 \leftarrow \#\{i: c_\theta(i){=}1,\, c_{\mathrm{mix}}(i){=}0\}$,\ \ $s_2 \leftarrow \#\{i: c_\theta(i){=}0,\, c_{\mathrm{mix}}(i){=}1\}$.
  \State Update $\alpha \leftarrow \sigma\bigl((s_2 - s_1 - c_o)/c_d\bigr)$ (Eq.~\eqref{eq:adapt-update}). 
  \Comment{\emph{with edge-case override; details below}}
\EndFor
\State \textbf{Output} $\pi_{\theta_{\mathrm{final}}}$ and $\pi^{\mathrm{mix}}_{\theta_{\mathrm{final}},\alpha}$.
\end{algorithmic}
\end{algorithm}

\textbf{An edge case and the fix}: Algorithm~\ref{alg:adaptive} can become ineffective if gradient updates induce unbounded growth in the magnitudes of the trainable logits. In this situation, they can overwhelm the frozen reference logits during averaging, dominating the mixed distribution regardless of the assigned weight. As a result, one could have $s_2 - s_1 - c_o\approx 0$ even if the reference anchor remains more accurate. This blind spot can lead to model collapse, preventing the mixed policy from effectively utilizing the reference policy. A fix to this edge case is to evaluate the trainable, mixed, and reference policies on the same held-out set. In particular, we introduce a conditional override in the validation loop: if $s_2 - s_1 - c_o\approx 0$ and $c_{\mathrm{ref}} > c_\theta$, we override the net-gain metric and set $\alpha \leftarrow 1$. Computationally, this fix introduces negligible overhead. Because the reference policy remains frozen, its validation can be performed once prior to the outer training loop and cached.

Throughout the rest of the paper, we use the SFT policy ($\pi_{\mathrm{sft}}$) as the reference policy ($\pi_{\mathrm{ref}}$).



\section{Averaging composes complementary skills}
\label{sec:mechanism}

Intuitively, when $\pi_{\text{sft}}$ and $\pi_\theta$ have complementary skills—for example, as illustrated in this section where $\pi_{\text{sft}}$ is strong at producing well-formatted responses (the \textit{format skill}) while $\pi_{\theta_{\text{final}}}$ is strong at reaching the correct numerical answer (the \textit{reasoning skill})—mixing their logits during post-training can take advantage of this complementarity. While we use stylistic formatting here as an illustrative example, the broader utility of LAPO lies in its ability to seamlessly integrate existing lightweight, parameter-efficient reference models. For instance, organizations can ensure adherence to strict corporate formatting rules or regulatory compliance laws (e.g., HIPAA, SEC guidelines) by anchoring their reasoning models with frozen specialized experts.


To examine the complementarity, for each problem, we record (i) whether the answer is enclosed in the required template (\emph{format-correct}) and (ii) simultaneous format correctness and numerical correctness (\emph{answer-correct}). We use the format-correct rate to measure the format skill and the answer-correct rate to measure the reasoning skill. The answer-correct rate is a \emph{conservative} estimate/lower bound for the reasoning skill: a response whose final number is correct but whose format is wrong is \emph{not} counted as answer-correct. 
Because $\pi_{\mathrm{sft}}$ has the stronger formatting ability, this underestimation is comparatively smaller for $\pi_{\mathrm{sft}}$ than for $\pi_{\theta_{\mathrm{final}}}$.


For each combination of a dataset in $\{\textrm{MATH},\textrm{MMLU},\textrm{cn-k12}\}$ and a model size in $\{\textrm{1.5B},\textrm{3B},\textrm{7B}\}$, we evaluate $\pi_{\theta_{\mathrm{final}}}$ from Algorithm~\ref{alg:fixed}  and $\pi_{\mathrm{sft}}$ separately on a held-out set of $N=200$ problems.
Aggregating across the 200 problems gives, for each $(\textrm{dataset},\textrm{model size})$ cell, a format-correct rate and an answer-correct rate across three policies: the frozen SFT policy $\pi_{\mathrm{sft}}$, the final trainable policy $\pi_{\theta_{\mathrm{final}}}$, and the equally weighted logit-averaging policy $\pi_{\theta_{\mathrm{mix}}}$. Table~\ref{tab:decoupling} reports all three comparisons, with the highest format-correct rate and highest answer-correct rate in each row shown in bold.


\paragraph{Observation 1: $\pi_{\mathrm{sft}}$ is better at format, $\pi_{\theta_{\mathrm{final}}}$ is better at reasoning.} Table~\ref{tab:decoupling} shows $\pi_{\mathrm{sft}}$ has a higher format-correct rate than $\pi_{\theta_{\mathrm{final}}}$ in $8$ of $9$ cells
. On the opposite, $\pi_{\theta_{\mathrm{final}}}$ outperforms $\pi_{\mathrm{sft}}$ in $8$ of $9$ cells for answer correctness, suggesting that 
during the training of Algorithm~\ref{alg:fixed}, the policy gradually achieves correct reasoning that the SFT checkpoint could not reach. Table~\ref{tab:gap} summarizes the per-cell correctness gap on each axis: the column ``Fmt: $\pi_{\mathrm{sft}}-\pi_{\theta_{\mathrm{final}}}$'' is positive when $\pi_{\mathrm{sft}}$ is the better formatter, and the column ``Ans: $\pi_{\theta_{\mathrm{final}}}-\pi_{\mathrm{sft}}$'' is positive when $\pi_{\theta_{\mathrm{final}}}$ is the better reasoner. With two exceptions, both columns are positive in every cell; that is, by construction of the training procedure, each policy has a different skill set. Since the answer-correct rate is a conservative proxy for reasoning, the reported ``Ans'' gap may carry some measurement error but remains effectively a lower bound for the true reasoning advantage of $\pi_{\theta_{\mathrm{final}}}$.




\paragraph{Observation 2: Logit averaging composes complementary skills.} No single policy dominates across all nine settings. SFT usually shows stronger formatting, whereas $\pi_{\theta_{\mathrm{final}}}$ usually improves answer correctness. The highest answer correctness in every setting is achieved by either $\pi_{\theta_{\mathrm{final}}}$ or $\pi_{\theta_{\mathrm{mix}}}$: $\pi_{\theta_{\mathrm{mix}}}$ is the best in five out of the nine settings, while $\pi_{\theta_{\mathrm{final}}}$ is the best in the remaining four. This observation supports the idea that logit averaging composes the complementary formatting and reasoning strengths of $\pi_{\mathrm{sft}}$ and $\pi_{\theta_{\mathrm{final}}}$.

\paragraph{Representative example.} Figure~\ref{fig:mechanism} illustrates the complementarity. The problem asks for the radius of the
circle $x^2+8x+y^2-6y=0$ where the golden answer is $5$. SFT completes the square but
puts $36$ on the right-hand side instead of $16+9{=}25$ and boxes the wrong
answer $6$; 
$\pi_{\theta_{\mathrm{final}}}$ gives the right algebra, simplifies the question to
$(x+4)^2+(y-3)^2=25$, and correctly observes ``radius $=\sqrt{25}=5$'', but
never emits a boxed final answer. The mixed policy at $\alpha{=}0.5$ writes
the correct derivation and terminates with the boxed $5$.

\begin{figure}[h]
\centering
\begin{tikzpicture}[
  font=\small,
  bar/.style={draw=black!60, thin},
  tag/.style={font=\scriptsize, text=black!70},
  sub/.style={font=\scriptsize\itshape, text=black!55},
  arr/.style={-{Latex[length=2mm]}, thick},
]
\begin{scope}[local bounding box=sftbox, shift={(0,3.5)}]
  \node[font=\bfseries] at (1.55, 1.60) {$\pi_{\mathrm{sft}}$};
  \node[sub]            at (1.55, 1.22) {sharp on \emph{format}, fragile on algebra};
  \draw[bar] (0.20, 0) rectangle (0.70, 0.08);
  \draw[bar, fill=orange!70] (0.95, 0) rectangle (1.45, 0.80);
  \draw[bar] (1.70, 0) rectangle (2.20, 0.12);
  \draw[bar] (2.45, 0) rectangle (2.95, 0.10);
  \node[tag] at (0.45,-0.22) {$5$};
  \node[tag] at (1.20,-0.22) {$\boxed{6}$};
  \node[tag] at (1.95,-0.22) {$25$};
  \node[tag] at (2.70,-0.22) {$\ldots$};
\end{scope}

\begin{scope}[local bounding box=rlbox, shift={(0,0)}]
  \node[font=\bfseries] at (1.55, 1.60) {$\pi_{\theta_{\mathrm{final}}}$};
  \node[sub]            at (1.55, 1.22) {sharp on \emph{algebra}, blurry on closing};
  \draw[bar, fill=blue!60] (0.20, 0) rectangle (0.70, 0.72);
  \draw[bar] (0.95, 0) rectangle (1.45, 0.10);
  \draw[bar] (1.70, 0) rectangle (2.20, 0.32);
  \draw[bar] (2.45, 0) rectangle (2.95, 0.22);
  \node[tag] at (0.45,-0.22) {$5$};
  \node[tag] at (1.20,-0.22) {$\boxed{6}$};
  \node[tag] at (1.95,-0.22) {$25$};
  \node[tag] at (2.70,-0.22) {$\ldots$};
\end{scope}

\begin{scope}[local bounding box=mxbox, shift={(7.5,1.85)}]
  \node[font=\bfseries] at (1.80, 2.10) {Mixed policy};
  \node[sub]            at (1.80, 1.72) {$\mathrm{softmax}\bigl((1{-}\alpha)z_\theta + \alpha z_{\mathrm{sft}}\bigr)$};
  \node[sub]            at (1.80, 1.35) {mass concentrates on tokens favoured by \emph{both} experts};
  \draw[bar] (0.05, 0) rectangle (0.55, 0.10);
  \draw[bar] (0.80, 0) rectangle (1.30, 0.14);
  \draw[bar, fill=green!65] (1.55, 0) rectangle (2.05, 0.90);
  \draw[bar] (2.30, 0) rectangle (2.80, 0.10);
  \draw[bar] (3.05, 0) rectangle (3.55, 0.08);
  \node[tag] at (0.30,-0.22) {$5$};
  \node[tag] at (1.05,-0.22) {$\boxed{6}$};
  \node[tag] at (1.80,-0.24) {$\boxed{5}$};
  \node[tag] at (2.55,-0.22) {$25$};
  \node[tag] at (3.30,-0.22) {$\ldots$};
\end{scope}

\begin{scope}[on background layer]
  \draw[rounded corners=4pt, thick, fill=orange!8]
    ($(sftbox.south west)+(-0.25,-0.50)$) rectangle ($(sftbox.north east)+(0.25,0.25)$);
  \draw[rounded corners=4pt, thick, fill=blue!8]
    ($(rlbox.south west)+(-0.25,-0.50)$)  rectangle ($(rlbox.north east)+(0.25,0.25)$);
  \draw[rounded corners=4pt, very thick, fill=green!6]
    ($(mxbox.south west)+(-0.25,-0.50)$)  rectangle ($(mxbox.north east)+(0.25,0.25)$);
\end{scope}

\draw[arr, orange!80!black]
  ($(sftbox.east)+(0.30,0)$) to[out=0,in=180] ($(mxbox.west)+(-0.30,0.55)$);
\draw[arr, blue!80!black]
  ($(rlbox.east)+(0.30,0)$)  to[out=0,in=180] ($(mxbox.west)+(-0.30,-0.55)$);


\end{tikzpicture}
\caption{
Mixing corrects SFT in numerical answer and improves upon RL in formatting.} 
\label{fig:mechanism}
\end{figure}

\section{Experiments}
\label{sec:empirical}
We evaluate the \emph{canonical} GRPO (with the KL regularization parameter $\beta{=}0.01$) in the form of equation (3) in \cite{grpo}, LAPO with fixed mixing (Algorithm~\ref{alg:fixed}, $\alpha{=}0.5$), and LAPO with adaptive mixing (Algorithm~\ref{alg:adaptive}) across three datasets: MATH (\texttt{Maxwell-Jia/MATH}), cn\_k12 (NuminaMath-1.5 source=\texttt{cn\_k12}), and MMLU. Qwen2.5-Instruct at 1.5B, 3B, and 7B serves our base policy and SFT serves our reference policy. We report greedy-decode accuracy of the \emph{mixed} policy (instead of the trainable policy) on a held-out set of $500$ problems.\footnote{Throughout all figures in this paper, one ``step'' on the horizontal axis denotes one outer iteration in LAPO concerning the index $t$.}



\subsection{
GRPO vs.\ Fixed-mixing LAPO}
\label{sec:emp-fixed-vs-grpo}

The fixed-mixing LAPO consistently outperforms the canonical KL-regularized GRPO, achieving a 7\%--15\% accuracy gain on MATH and a 2\%--5\% gain on cn\_k12 and MMLU
(Figures~\ref{fig:emp-math-fixed}, \ref{fig:emp-cnk12-fixed} and \ref{fig:emp-mmlu-fixed} in Appendix \ref{appedix:supporting figures}). The KL-regularized GRPO here is initialized from $\pi_{\mathrm{sft}}$, following the standard SFT-then-RL recipe. The fixed-mixing LAPO also outperforms GRPO with $\beta=0$  and initialization $\pi_{\mathrm{sft}}$ (Figure~\ref{fig:emp-math-fixed-vs-grpo0} in Appendix \ref{appedix:supporting figures}). This result suggests that 
even though initialized from $\pi_{\mathrm{sft}}$, the trainable policy drifts away from $\pi_{\mathrm{sft}}$ during training and forgets the information in $\pi_{\mathrm{sft}}$, whereas the fixed-mixing LAPO preserves this information by directly incorporating $\pi_{\mathrm{sft}}$ into the rollout policy at the logit level throughout training.








\subsection{Fixed-mixing LAPO vs.\ Adaptive-mixing LAPO}
\label{sec:emp-fixed-vs-adaptive}

On MATH, the fixed-mixing LAPO and the adaptive-mixing LAPO show comparable accuracy in Figures \ref{fig:emp-math-adaptive} and \ref{fig:emp-mmlu-adaptive} (in Appendix \ref{appedix:supporting figures}). We observe that the fixed-mixing and adaptive-mixing LAPO obtain almost similar results at the end. On MMLU 3B, the adaptive-mixing LAPO ends $1$--$2$ basis points above the fixed-mixing counterpart. We think the comparable performance of the two algorithms may be due the dominance of $\pi_\theta$ after enough gradient updates of $\theta$ in the datasets we study. It is possible that adaptive-mixing LAPO may outperform the fixed-mixing counterpart in other datasets.

\subsection{Anchor ablation: SFT vs. base}
\label{sec:emp-anchor}

To better understand the role of SFT in logit averaging, we rerun the fixed-mixing LAPO with the base Instruct model replacing the SFT anchor on cn\_k12 (Figure~\ref{fig:emp-cnk12-base} in Appendix \ref{appedix:supporting figures}). Starting much lower, the base-anchor variant never quite catches up to the SFT-anchor variant. This finding suggests that the gains we report earlier are from mixing with a \emph{format-aware} reference distribution.


\subsection{Logit averaging vs. probability averaging}
\label{sec:numerical}

An obvious alternative is to average two policies in the probability space: $\pi^{\mathrm{prob}}_{\theta,\alpha}\!=\!(1{-}\alpha)\pi_{\theta}+\alpha\pi_{\mathrm{ref}}$, which corresponds to an \emph{arithmetic} mean. In contrast, logit averaging is related to a (re-normalized) \emph{geometric} mean of two predictive distributions (see detailed discussions in Section~\ref{sec:discussion}). 

To compare the two different mixing schemes, we evaluate them in Qwen2.5-1.5B-Instruct on the cn\_k12 held-out set (500 problems) with $\alpha\in\{0.0,0.1,\dots,1.0\}$, and Table~\ref{tab:logit-vs-prob} reports the format rate (whether the response follows the required answer template) and solve rate (whether both the format and final answer are correct) at inference time. In addition, Figure~\ref{fig:logit-vs-prob-training} plots the 
MATH solve rate of $\pi^{\mathrm{mix}}_{\theta,\alpha}$ across $40$ 
steps in Algorithm~\ref{alg:fixed} for Qwen2.5-3B-Instruct 
based on logit mixing $z^{\mathrm{mix}}_{\theta,0.5}=\tfrac{1}{2}z_\theta+\tfrac{1}{2}z_{\mathrm{ref}}$ and probability mixing $\pi^{\mathrm{prob}}_{\theta,0.5}=\tfrac{1}{2}\pi_\theta+\tfrac{1}{2}\pi_{\mathrm{ref}}$, respectively. Due to space constraints, the discussion of these results is deferred to Appendix \ref{appedix:supporting figures}.




\section{Discussions}
\label{sec:discussion}


We first provide three complementary views on logit averaging. 

\textbf{Connection to shrinkage estimators and model averaging.}\hspace{1em}From a classical statistical perspective, model averaging and parameter shrinkage (e.g., Ridge regression with an $l_2$-penalty) are deeply connected mechanisms designed to reduce estimator variance and lower mean squared error. Rather than averaging high-dimensional weight parameters directly, logit averaging functions as an implicit structural regularizer in the logit space. 
When paired with adaptive mixing-weight selection (Algorithm \ref{alg:adaptive}), logit averaging helps stabilize RL exploration and prevent erratic policy drift—delivering the benefits of regularization without relying on KL penalties and expensive hyperparameter sweeps.

\textbf{Connection to Products of Experts.}\hspace{1em}Given a state $s$, let $\pi_1=\pi_\theta(\cdot|s)$ and  $\pi_2=\pi_{\mathrm{ref}}(\cdot|s)$
denote two distributions on a finite action set $\mathcal{A}$. We suppress $s$ below to simplify the notation.  
Let $z_i(a') = \log\pi_i(a') + c_i$ where $c_i=\log \sum_{a'}\exp(z_i(a'))$ for $i\in\{1,2\}$. 
Writing $z_{\mathrm{mix}}(a)=(1-\alpha)z_1(a)+\alpha z_2(a)$, we have 
\begin{align}
\label{eq:poe}
\mathrm{softmax}z_{\mathrm{mix}}(a)
&=
\arg\min_{p\in\Delta(\mathcal{A})}\Bigl[\,(1-\alpha)\,\mathrm{KL}(p\|\pi_1) + \alpha\,\mathrm{KL}(p\|\pi_2)\Bigr]
&=
\frac{\pi_1(a)^{1-\alpha}\,\pi_2(a)^{\alpha}}{\sum_{a'} \pi_1(a')^{1-\alpha}\pi_2(a')^{\alpha}}.
\end{align}
At each state, a softmax operation on the logit averaging yields the policy that is simultaneously closest to the RL policy and the reference policy in
reverse KL, weighted by $1-\alpha$ and $\alpha$ respectively. Eq.~\eqref{eq:poe} takes the form of a \emph{Product of Experts} (PoE)
\citep{hinton2002poe}.

\textbf{Connection to the SDPO trust-region teacher.}\hspace{1em}Eq.~\eqref{eq:poe} takes a form similar to
the trust-region regularized teacher derived in Section~B.2 of
\citet{sdpo2026}. In the notation of \citet{sdpo2026}, their Proposition~B.1 shows that the optimal teacher policy $q^\star$ has minimal cross-entropy to $q_\theta(y_t\mid x,f,y_{<t})$ and maximal cross-entropy to $q_{\theta_{\mathrm{ref}}}(y_t\mid x,f,y_{<t})$ while satisfying a trust region type of constraint as in \citep{schulman2015trpo,peng2019awr}; in particular, 
\begin{align}
\label{eq:teacher EMA}
q^\star(y_t\mid x,f,y_{<t}) \;\propto\; \exp\bigl((1-\alpha)\log q_{\theta_{\mathrm{ref}}}(y_t\mid x,f,y_{<t}) + \alpha\log q_\theta(y_t\mid x,f,y_{<t})\bigr)
\end{align}
where $f$ in \citet{sdpo2026} refers to feedback and $\alpha$ is the inverse of the Lagrange multiplier from the Lagrange dual formulation of the trust region constraint. 

\textbf{What differentiate our contributions?}\hspace{1em}\citet{sdpo2026} minimizes a logit distillation loss 
\begin{equation*}
\mathcal{L}_{\textrm{SDPO}}(\theta)=\sum_t \mathrm{KL}(\pi_\theta(\cdot \vert x,y_{<t})\|\ \mathrm{stopgrad}(\pi_\theta(\cdot \vert x,f,y_{<t})))
\end{equation*}
and computes the advantages in the form of $A^{\textrm{SDPO}}_{i,t}(\hat{y}_{i,t})=\log \frac{\pi_\theta(\cdot \vert x,f,y_{<t}))}{\pi_\theta(\cdot \vert x,y_{<t})}$; in particular, $q^\star(y_t\mid x,f,y_{<t})$ in (\ref{eq:teacher EMA}) replaces $\pi_\theta(\cdot \vert x,f,y_{<t})$ for stability improvement in on-policy distillation from teachers but a similar strategy is not used on $\pi_\theta(\cdot \vert x,y_{<t})$. In contrast, as detailed in Section \ref{sec:method}, we employ a modified GRPO where the entire stack (e.g., rollout, policy ratio, and surrogate objective) is based on our mixed policies. In addition, our algorithms involve no extra information such as $f$ in \citet{sdpo2026} and \citet{shenfeld2026}, no new reward models, and no stronger teachers during training. As a result, any improvement over pure RL shown in our results is only
attributable to the mixing mechanism itself. This fact makes a more insightful comparison
between algorithms with a fixed data budget.

\textbf{Inference-time PoE vs. Training-time PoE.}\hspace{1em}In the LLM literature, the PoE idea has been used as an \emph{inference-time} controlled-decoding mechanism, where two language models are combined in the logit space at every decoding step during inference. DExperts \citep{liu2021dexperts} steers a base language model by adding the logit difference of an ``expert'' and ``anti-expert'' fine-tuned on desirable/undesirable text. Contrastive Decoding \citep{li2023contrastive} samples from $\log p_{\mathrm{expert}} - \log p_{\mathrm{amateur}}$ to suppress generic continuations of a smaller model. Proxy Tuning \citep{liu2024proxy} proposes a large base model with the logit-shift of a small tuned/base pair, so that a large model effectively inherits the behavior of a small fine-tuned one without ever updating its parameters. 
THINKMERGE \citep{wang2026thinkmerge} averages logits conditioned on multiple parallel reasoning traces during answer generation. DeRa \citep{liu2024dera} combines a frozen SFT reference policy with an aligned policy, allowing the effective alignment strength to be adjusted at decoding time without retraining. MOD \citep{shi2024mod} extends inference-time PoE to multiple objective-specific policies by combining their predictions according to user-specified preference weights. However, running multiple models at inference time increases inference latency and memory footprint. In contrast to this literature, we push this PoE structure \emph{inside the RL training loop}, effectively retaining SFT formatting rigor while empowering the trainable policy to reason more freely. Standard single-network efficiency can be restored by distilling the mixed teacher into a student model before deployment. Alternatively, evaluating whether the final trainable policy $\pi_{\theta_{\mathrm{final}}}$ alone sufficiently retains or even exceeds these performance gains without requiring reference model mixing remains another promising direction for future investigation.

\textbf{Logit averaging vs KL regularization.}\hspace{1em}As discussed in \citet{korbak2022rl},
any RL objective function with a $\mathrm{KL}(\pi_\theta\|\pi_{\mathrm{ref}})$ penalty admits the closed-form optimum $\pi^{\star}\propto \pi_{\mathrm{ref}}\exp(R/\beta)$; moreover, the objective can be rewritten as a KL minimization toward a target that contains $\pi_{\mathrm{ref}}$ as a multiplicative factor. Since gradient updates bring the policy close to the optimum, the resulting $\pi_\theta$ tends to inherit the support and shape of $\pi_{\mathrm{ref}}$, which may, in turn, transmit limitations of the reference policy back into $\pi_\theta$. Our training objective, by contrast, does not contain a divergence term that pulls $\pi_\theta$ toward a target shaped by $\pi_{\mathrm{sft}}$; the SFT anchor enters only through the geometric-mean mixed policy used during rollout and in the surrogate. This structure may give $\pi_\theta$ somewhat more freedom to compensate for $\pi_{\mathrm{sft}}$'s deficiencies. For example, the structure might encourage concentrating probability on tokens that $\pi_{\mathrm{sft}}$ under weighs while still leveraging the formatting prior that $\pi_{\mathrm{sft}}$ supplies through mixing. It is important emphasizing that the above discussion may not fully explain the performance gap observed in the data sets we study here or guarantee better performance of LAPO in data sets where the complementarity pattern discussed in Section \ref{sec:mechanism} does not hold. A deeper investigation is left to future work. 





\textbf{Why could logit averaging outperform pure RL?}\hspace{1em}We lay out two non-exclusive hypotheses. First, mixing the RL logits with the SFT logits is an implicit trust region strategy. This fact prevents the slow
distributional drift-away from formatting and style,
which is known to harm pure RL at scale. Second, the
geometric structure of logit averaging (Eq.~\eqref{eq:poe}) complements skills identified in Section~\ref{sec:mechanism} in a way that
the RL gradient alone cannot: RL propagates backward the signal from rewarding the final answer token, whereas logit averaging incorporates a dense token-level prior from the SFT anchor at every decoding step.

\textbf{Training cost comparison with GRPO}\hspace{1em}In post-training, removing KL regularization eliminates expensive, fragile hyperparameter sweeps and penalty autograd overhead. Most importantly, LAPO can improve sample efficiency by requiring fewer rollouts. Under binary rewards and finite group sizes ($G$), standard GRPO even with KL regularization frequently samples groups with zero successful responses---producing identical zero rewards and uninformative zero-advantages. Mixing the logit of the trainable policy with the logit of an SFT anchor that assigns higher prior probability to well-formed responses helps enrich the exploration distribution and substantially reduce the incidence of these uninformative zero-reward groups. Empirically, Figure~\ref{fig:math-group-size} demonstrates that \textbf{LAPO (Algorithm~\ref{alg:adaptive}) with $G=4$ achieves accuracy on par with $G=8$, effectively cutting rollout generation costs in half.} On the other hand, LAPO incurs higher peak activation memory from materializing full-vocabulary logit tensors during loss evaluation; mitigating this via fused Triton kernels or top-$k$ truncation ($k \ll V$) remains future work.\footnote{In addition, when the reference policy shares the base model with the trainable policy, employing LoRA \cite{hu2021lora} offers a natural path to reduce memory footprints.} 

\textbf{Leveraging small language models expertise.}\hspace{1em}Our method naturally extends to ensembles of Small Language Models (SLMs):
$z_{\text{mix}}(s) = \sum_{i=1}^M \beta_i z_{\theta, i}(s) + \sum_{j=1}^N \gamma_j z_{\text{ref}, j}(s)$,
where $z_{\theta, i}$ denotes the logits of the $i$-th trainable policy undergoing active post-training, $z_{\text{ref}, j}$ denotes the logits of the $j$-th frozen reference model specialized in a static capability, and mixing coefficients $\sum_i \beta_i + \sum_j \gamma_j = 1$. For example, one or more active reasoning policies can be updated via RL while anchored by lightweight, frozen experts fine-tuned for code syntax, domain-specific terminology (e.g., medical or legal), or safety guardrails. This structure allows the system to summon and scale SLM expertise while reducing the prohibitive cost of training a massive policy. This is particularly valuable during post-training in resource-constrained edge AI settings, where compute-intensive hyperparameter sweeps are infeasible.



\newpage
\appendix

{\LARGE\bfseries SUPPLEMENTARY MATERIALS: LAPO: KL-Free Logit-Averaged Policy Optimization
to Leverage RL and SFT in LLM Post-Training\par}
\vspace{0.5cm} 

\section{Tables, figures and additional discussions}
\label{app:additional}

\subsection{Supporting materials for Section~\ref{sec:mechanism}}

\begin{table}[h]
\centering
\caption{Format-correct rate and answer-correct rate  (in percent) on the same 200 held-out problems for SFT, $\theta_{\mathrm{final}}$, and $\theta_{\mathrm{mix}}$. Bold denotes the highest format-correct rate and highest answer-correct rate in each row. Recall that answer correctness means simultaneous format correctness and numerical correctness and the answer-correct rate is a \emph{conservative} estimate/lower bound for the reasoning skill.}
\label{tab:decoupling}
\resizebox{\textwidth}{!}{%
\begin{tabular}{llccc}
\toprule
Dataset & Model & SFT Format / Ans & $\theta_{\mathrm{final}}$ Format / Ans & $\theta_{\mathrm{mix}}^{\dagger}$ Format / Ans \\
\midrule
MATH   & 1.5B & $91.0/39.5$ & $63.0/53.0$ & $\mathbf{92.0}/\mathbf{67.5}$ \\
MATH   & 3B   & $\mathbf{95.0}/59.0$ & $90.5/76.5$ & $91.5/\mathbf{79.0}$ \\
MATH   & 7B   & $\mathbf{96.5}/68.5$ & $94.5/88.5$ & $\mathbf{96.5}/\mathbf{89.0}$ \\
MMLU   & 1.5B & $\mathbf{100}/40.0$ & $94.5/\mathbf{44.0}$ & $\mathbf{100}/42.5$ \\
MMLU   & 3B   & $\mathbf{100}/48.0$ & $99.0/\mathbf{57.5}$ & $99.0/52.5$ \\
MMLU   & 7B   & $\mathbf{100}/64.0$ & $98.5/\mathbf{71.0}$ & $99.5/68.0$ \\
cn-k12 & 1.5B & $\mathbf{81.5}/27.5$ & $28.5/21.0$ & $70.5/\mathbf{43.0}$ \\
cn-k12 & 3B   & $\mathbf{87.0}/37.5$ & $78.5/56.0$ & $83.0/\mathbf{57.5}$ \\
cn-k12 & 7B   & $84.0/58.5$ & $\mathbf{92.5}/\mathbf{68.5}$ & $87.0/66.0$ \\
\bottomrule
\end{tabular}%
}
\begin{minipage}{\textwidth}
\footnotesize
$^{\dagger}\theta_{\mathrm{mix}}$ denotes the policy obtained by equally weighting the logits of the final trainable policy and the frozen SFT policy: $\pi_{\theta_{\mathrm{mix}}}=\operatorname{softmax}(0.5z_{\theta_{\mathrm{final}}}+0.5z_{\mathrm{sft}})$.
\end{minipage}
\end{table}

\begin{table}[h]
\centering
\caption{Skill difference of $\pi_{\mathrm{sft}}$ and $\pi_{\theta_{\mathrm{final}}}$}
\label{tab:gap}
\begin{tabular}{llrr}
\toprule
Dataset & Model & Fmt: $\pi_{\mathrm{sft}}-\pi_{\theta_{\mathrm{final}}}$ (pp) & Ans: $\pi_{\theta_{\mathrm{final}}}-\pi_{\mathrm{sft}}$ (pp) \\
\midrule
MATH   & 1.5B & $+28.0$ & $+13.5$ \\
MATH   & 3B   & $+4.5$  & $+17.5$ \\
MATH   & 7B   & $+2.0$  & $+20.0$ \\
MMLU   & 1.5B & $+5.5$  & $+4.0$  \\
MMLU   & 3B   & $+1.0$  & $+9.5$  \\
MMLU   & 7B   & $+1.5$  & $+7.0$  \\
cn-k12 & 1.5B & $+53.0$ & $-6.5$  \\
cn-k12 & 3B   & $+8.5$  & $+18.5$ \\
cn-k12 & 7B   & $-8.5$  & $+10.0$ \\
\bottomrule
\end{tabular}
\end{table}

\subsection{Supporting materials for Sections~\ref{sec:empirical} and ~\ref{sec:discussion}}\label{appedix:supporting figures} 

\begin{figure}[h]
\centering
\includegraphics[width=0.32\linewidth]{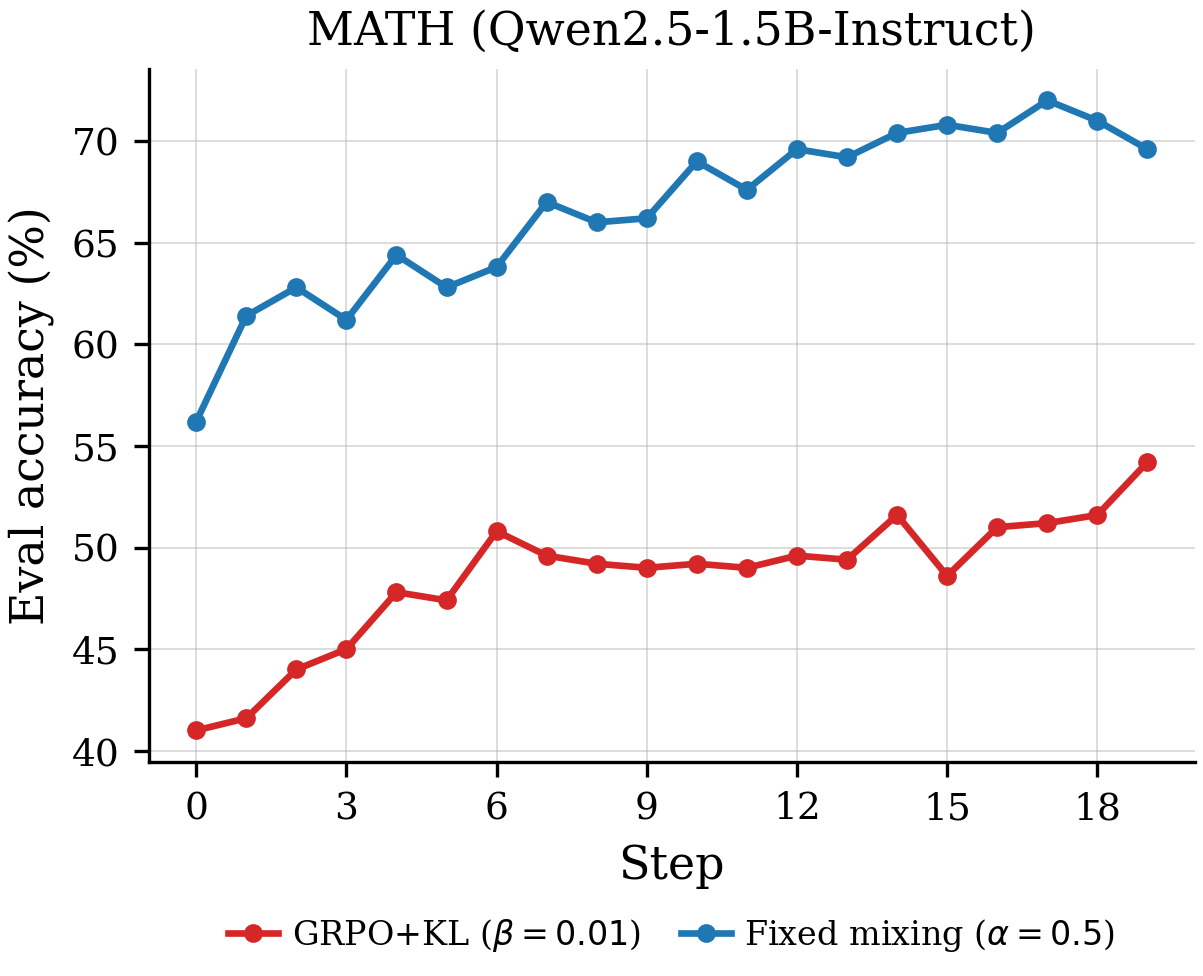}\hfill
\includegraphics[width=0.32\linewidth]{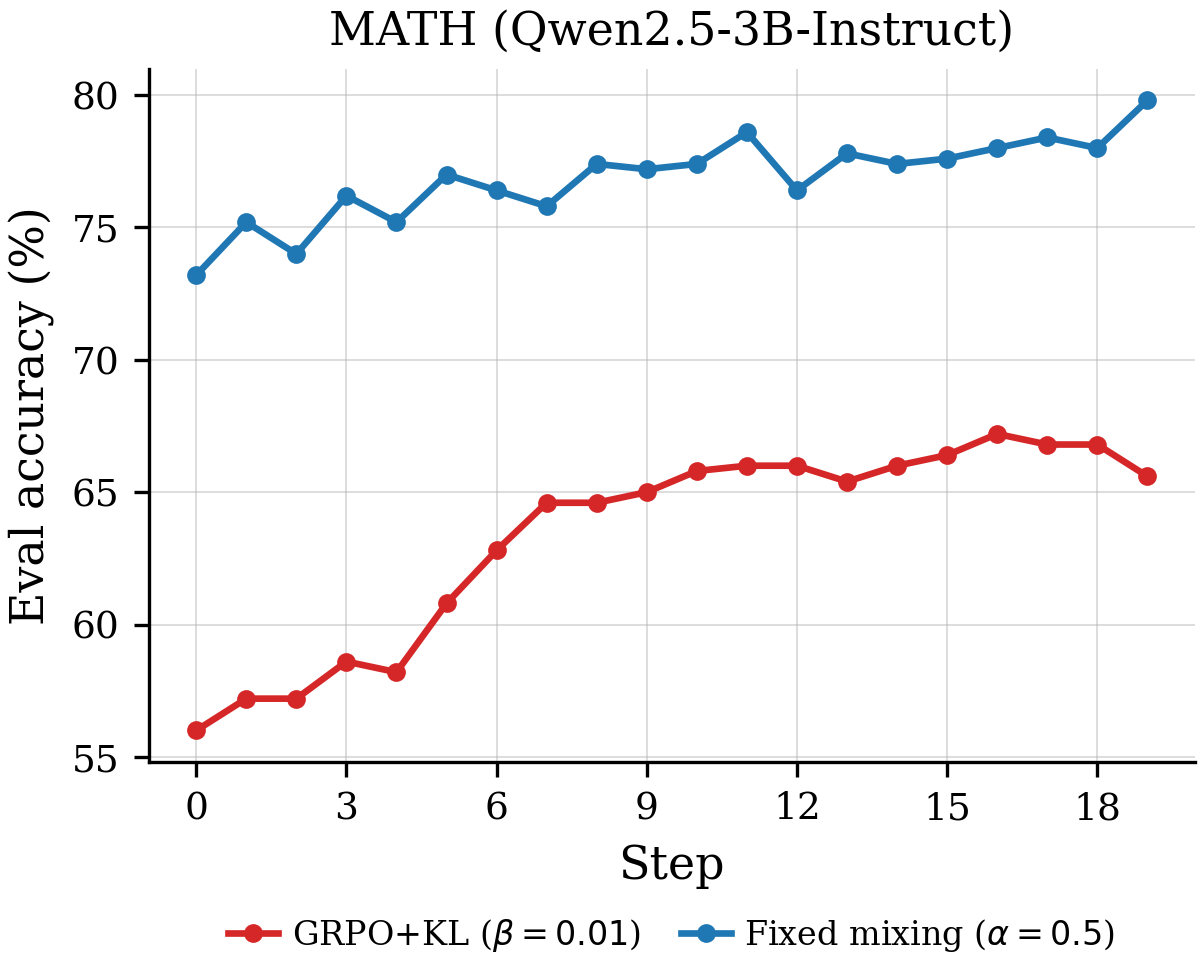}\hfill
\includegraphics[width=0.32\linewidth]{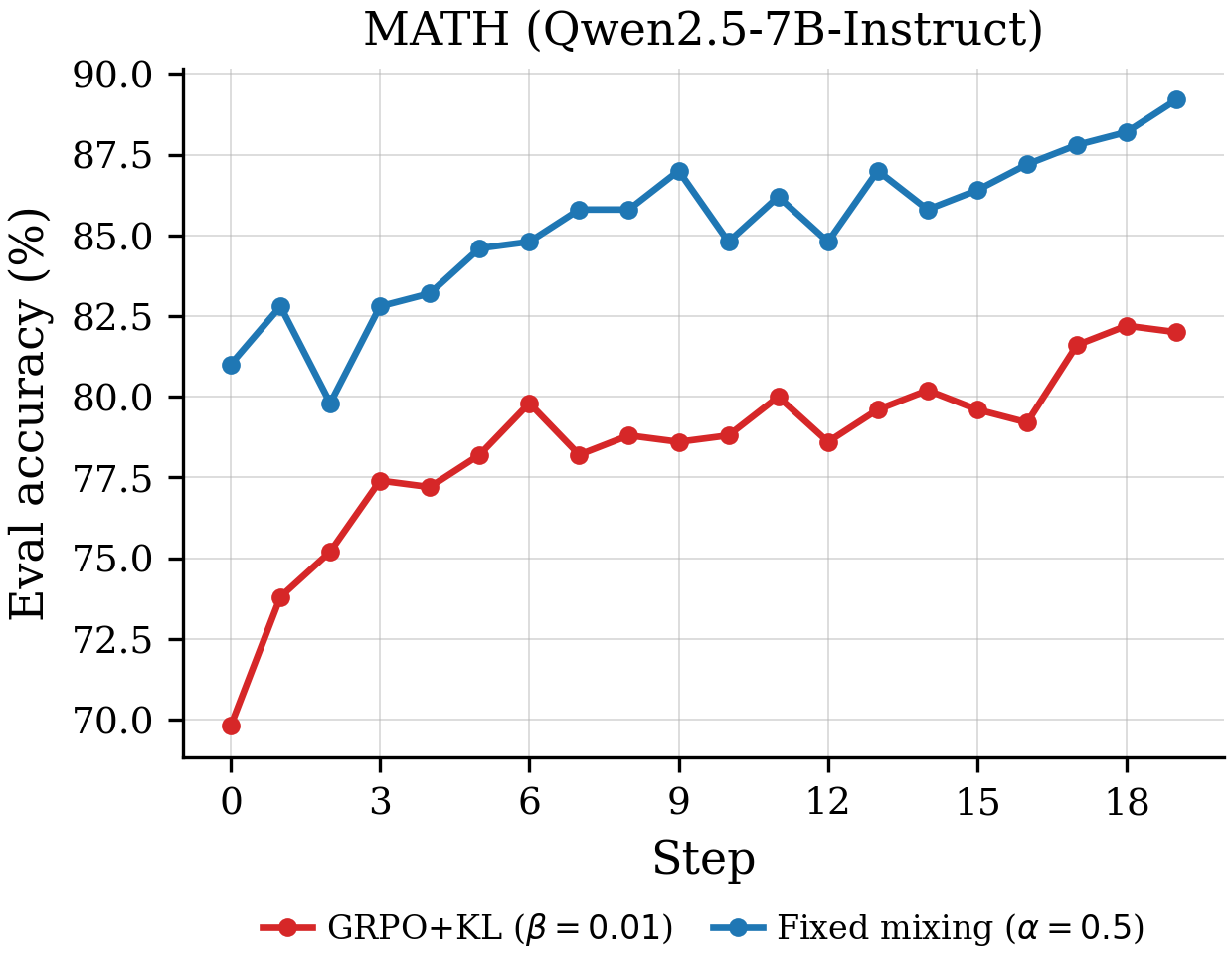}
\caption{MATH: KL-regularized GRPO vs.\ fixed-mixing LAPO across 1.5B, 3B, 7B Qwen2.5-Instruct policies.}
\label{fig:emp-math-fixed}
\end{figure}

\begin{figure}[h]
\centering
\includegraphics[width=0.55\linewidth]{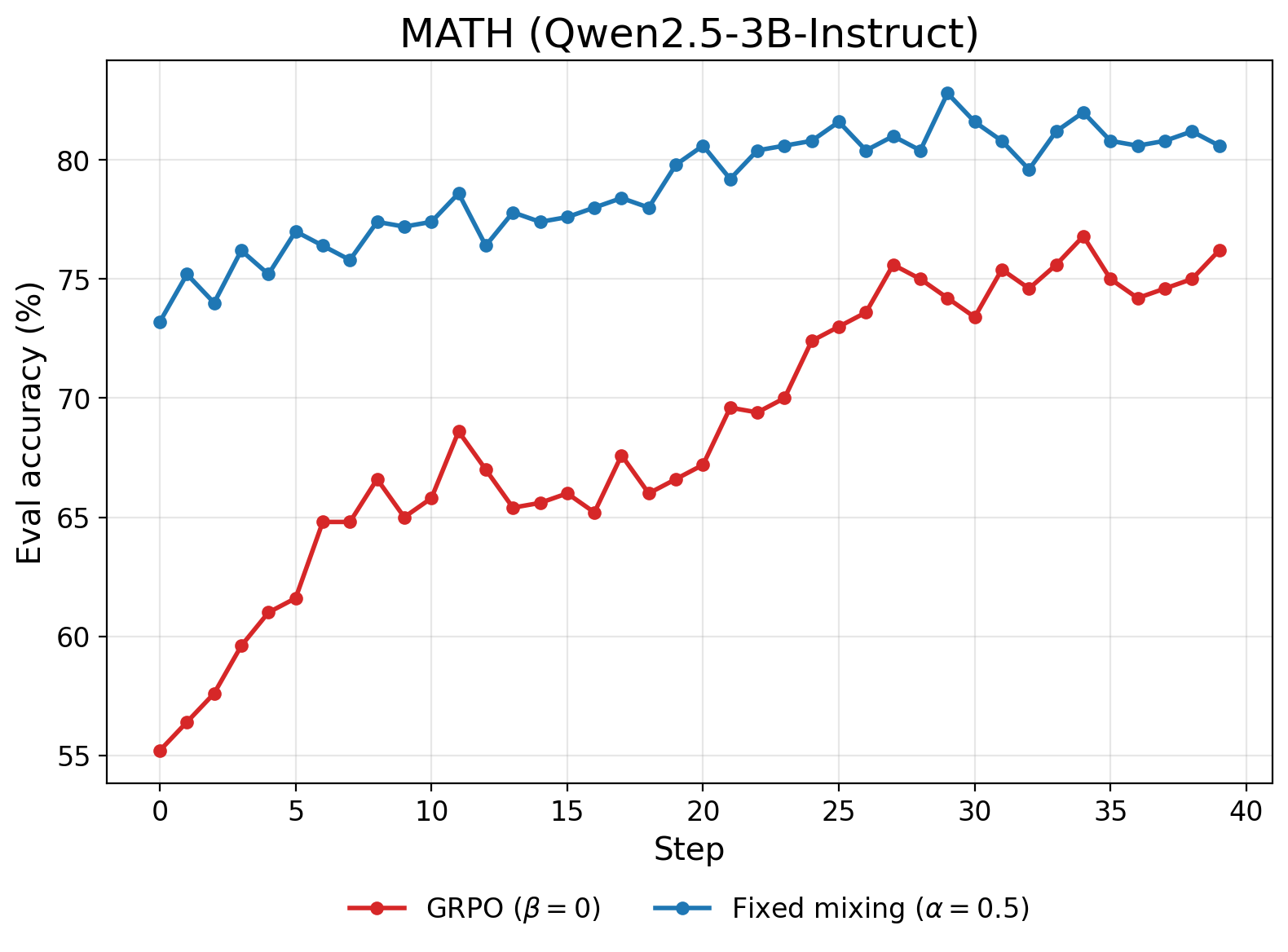}
\caption{MATH (Qwen2.5-3B-Instruct): Fixed-mixing LAPO ($\alpha=0.5$) vs.\ GRPO with $\beta=0$ (no KL regularization).}
\label{fig:emp-math-fixed-vs-grpo0}
\end{figure}

\begin{figure}[h]
\centering
\includegraphics[width=0.32\linewidth]{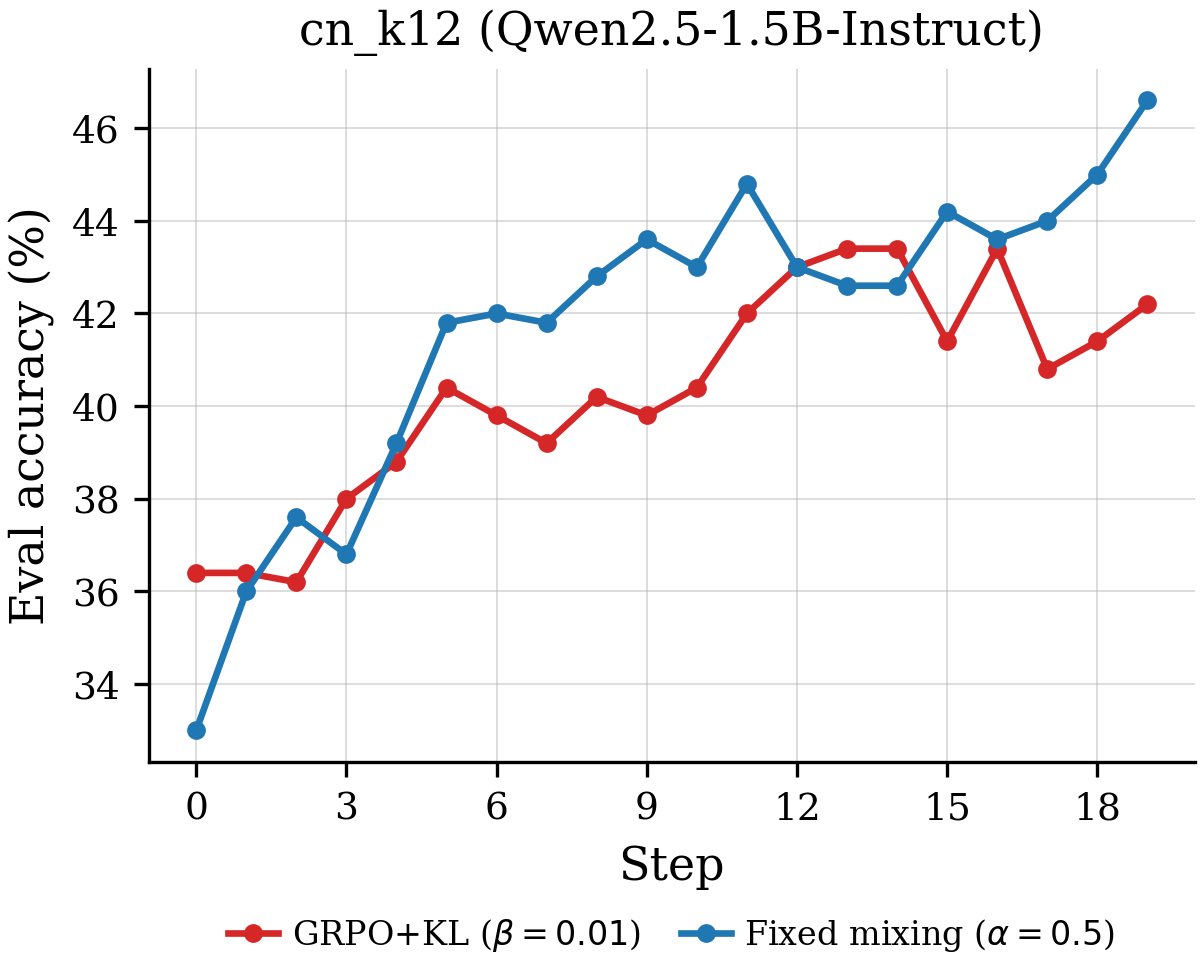}\hfill
\includegraphics[width=0.32\linewidth]{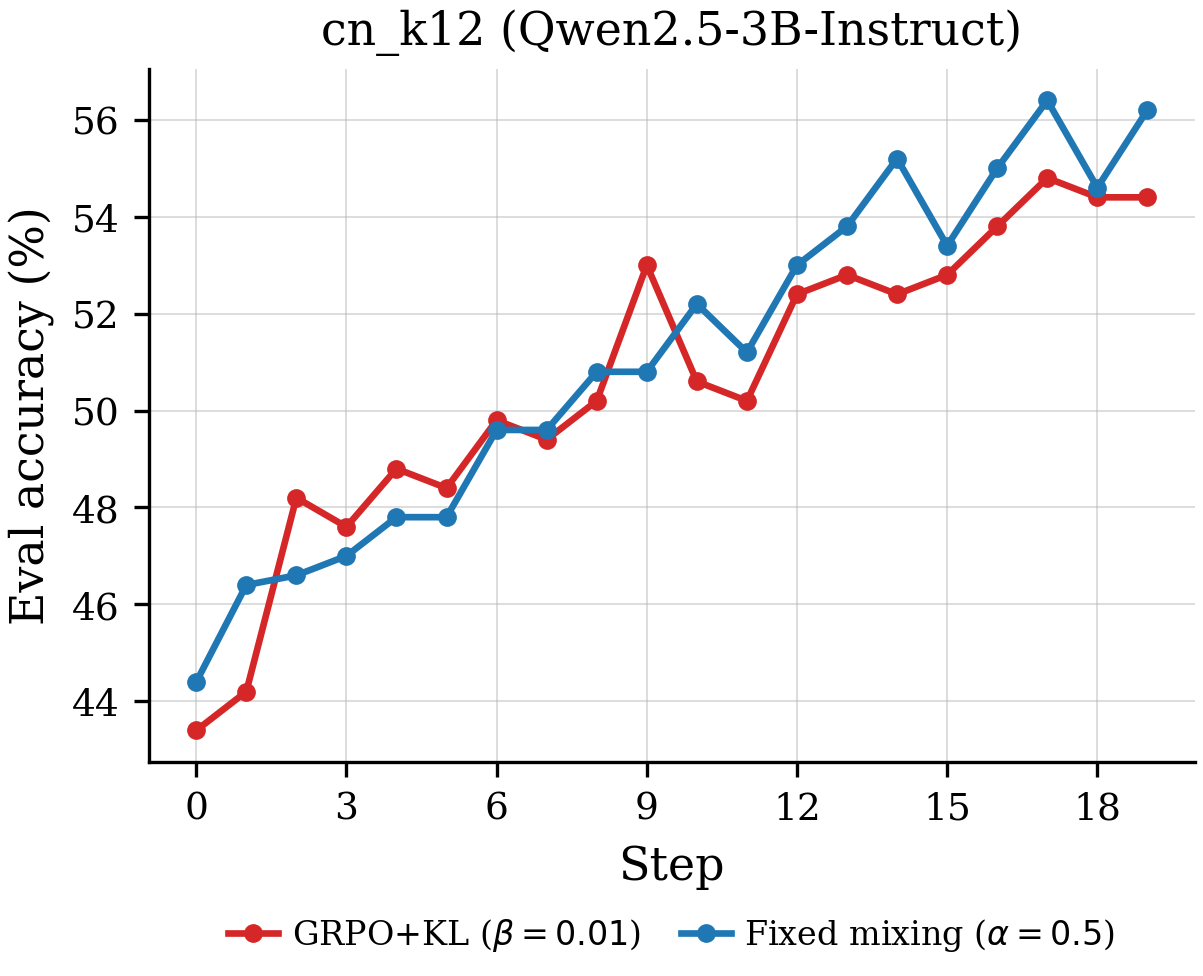}\hfill
\includegraphics[width=0.32\linewidth]{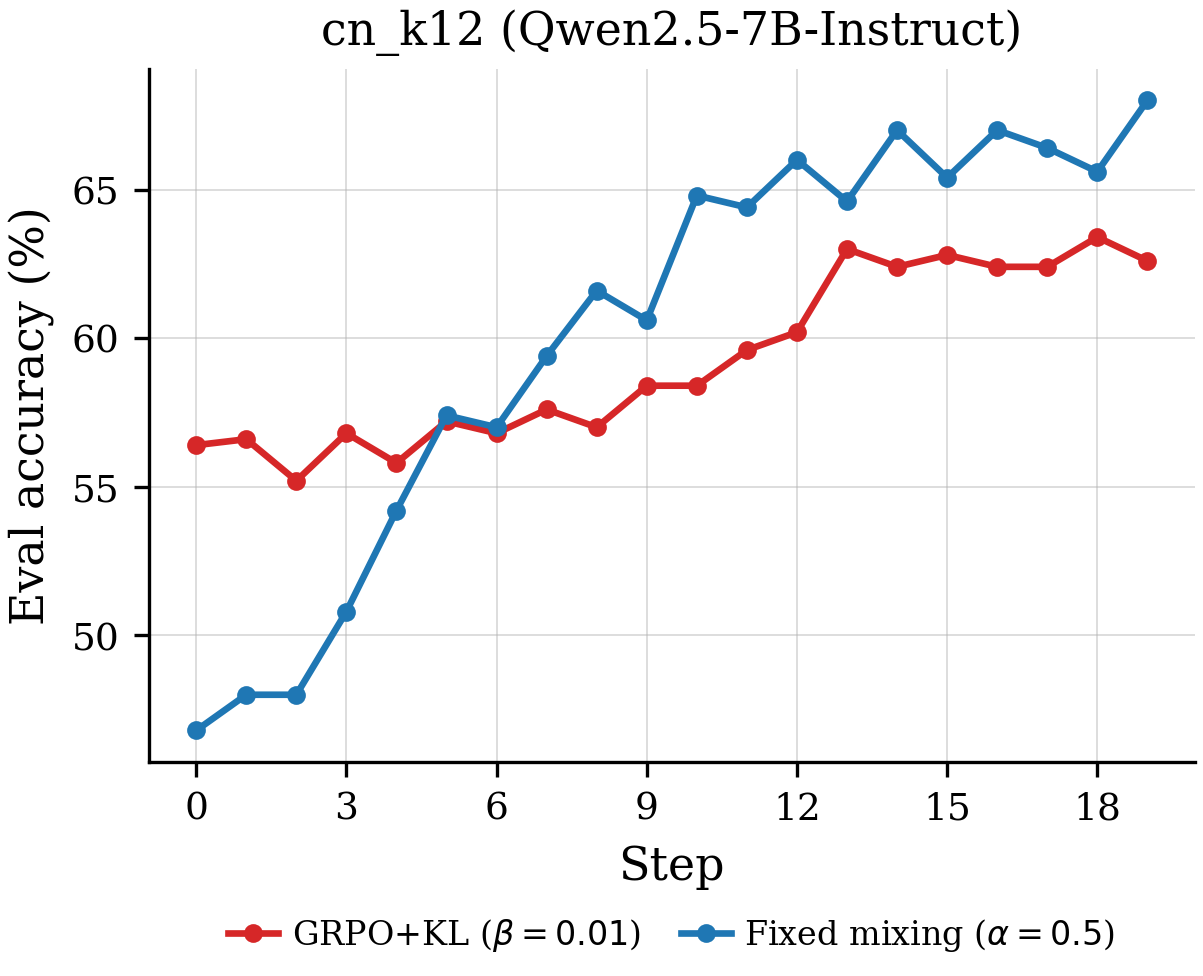}
\caption{cn\_k12: KL-regularized GRPO vs.\ fixed-mixing LAPO across 1.5B, 3B, 7B Qwen2.5-Instruct policies.}
\label{fig:emp-cnk12-fixed}
\end{figure}

\begin{figure}[h]
\centering
\includegraphics[width=0.32\linewidth]{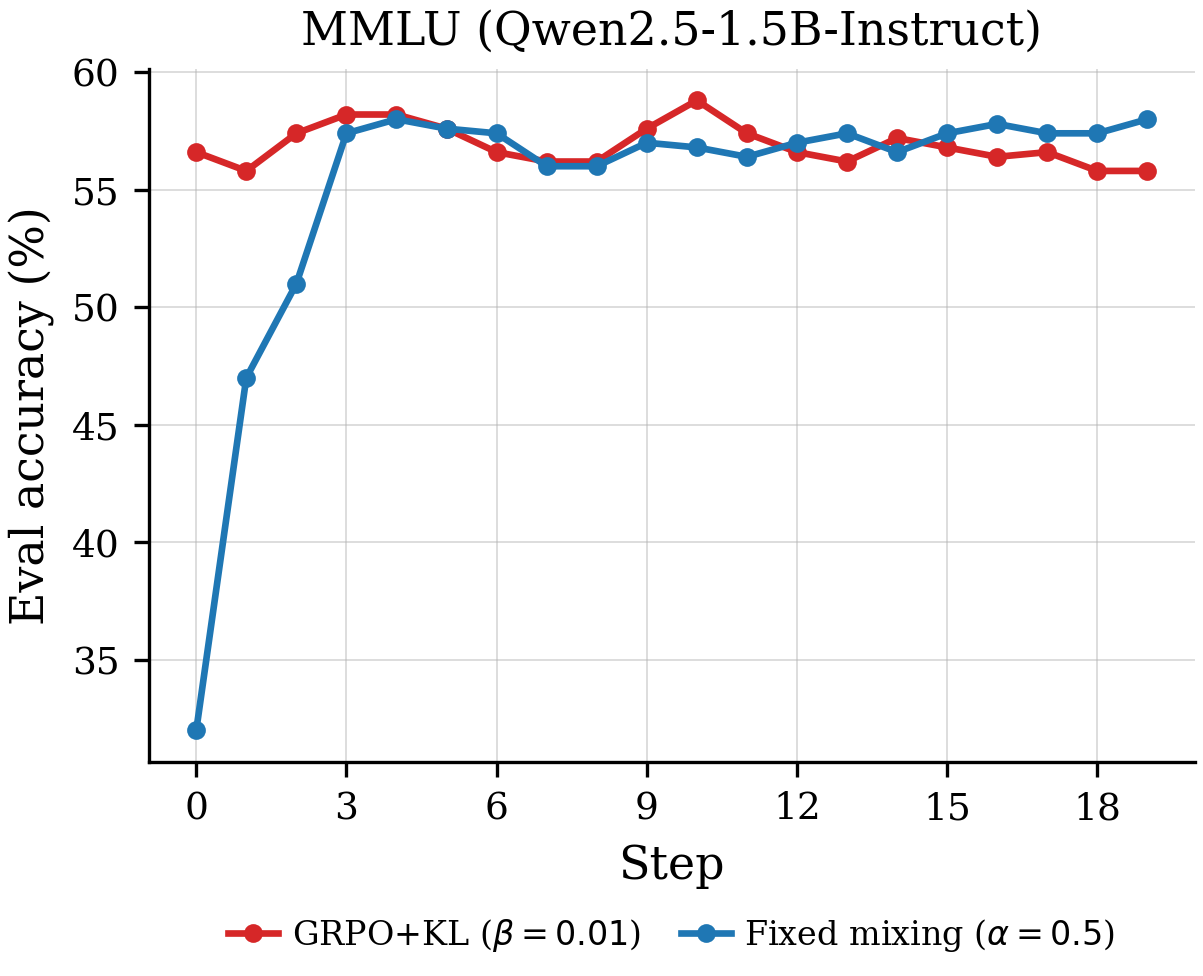}\hfill
\includegraphics[width=0.32\linewidth]{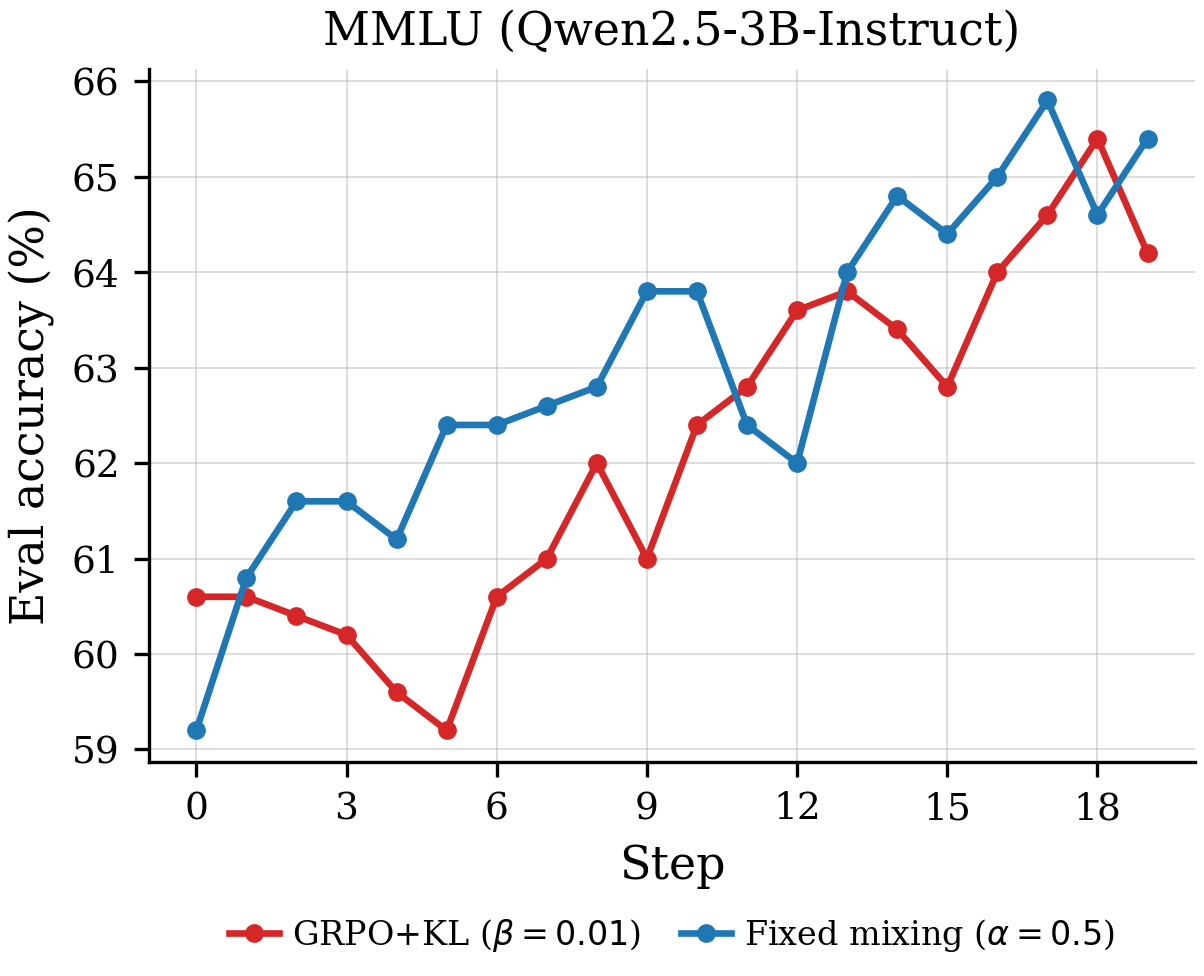}\hfill
\includegraphics[width=0.32\linewidth]{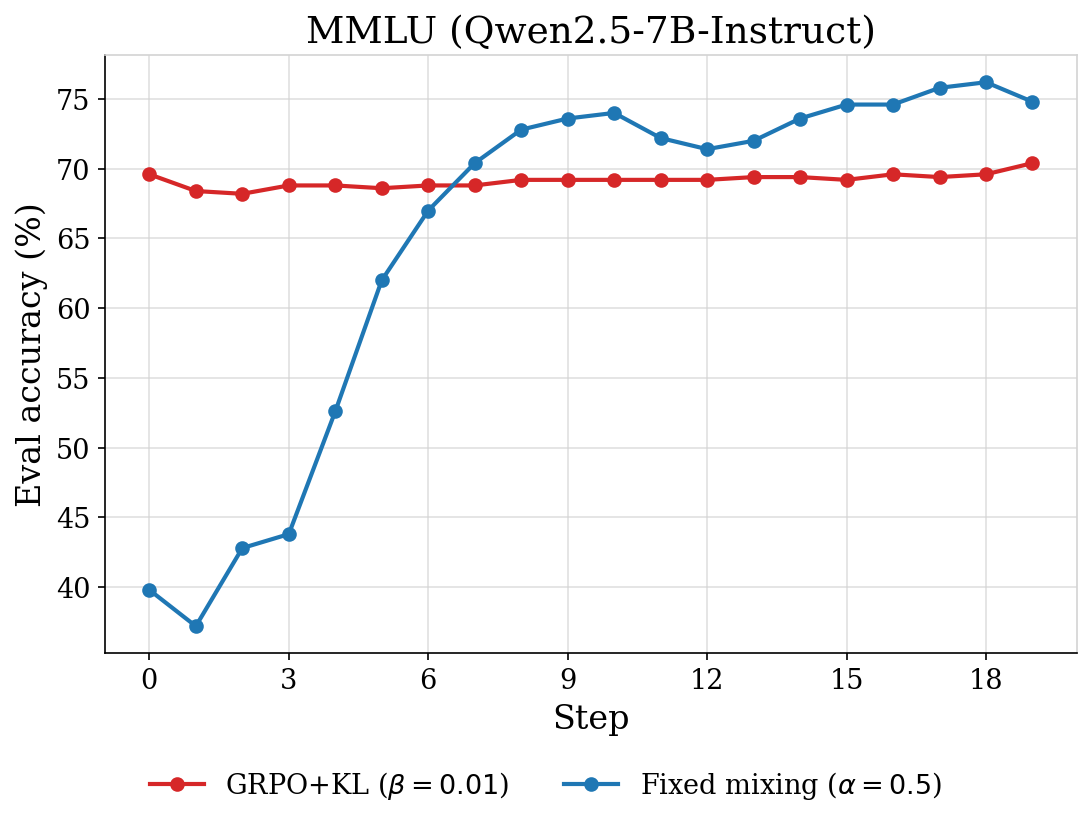}
\caption{MMLU: KL-regularized GRPO vs.\ fixed-mixing LAPO at 1.5B, 3B and 7B.}
\label{fig:emp-mmlu-fixed}
\end{figure}




\begin{figure}[h]
\centering
\includegraphics[width=0.32\linewidth]{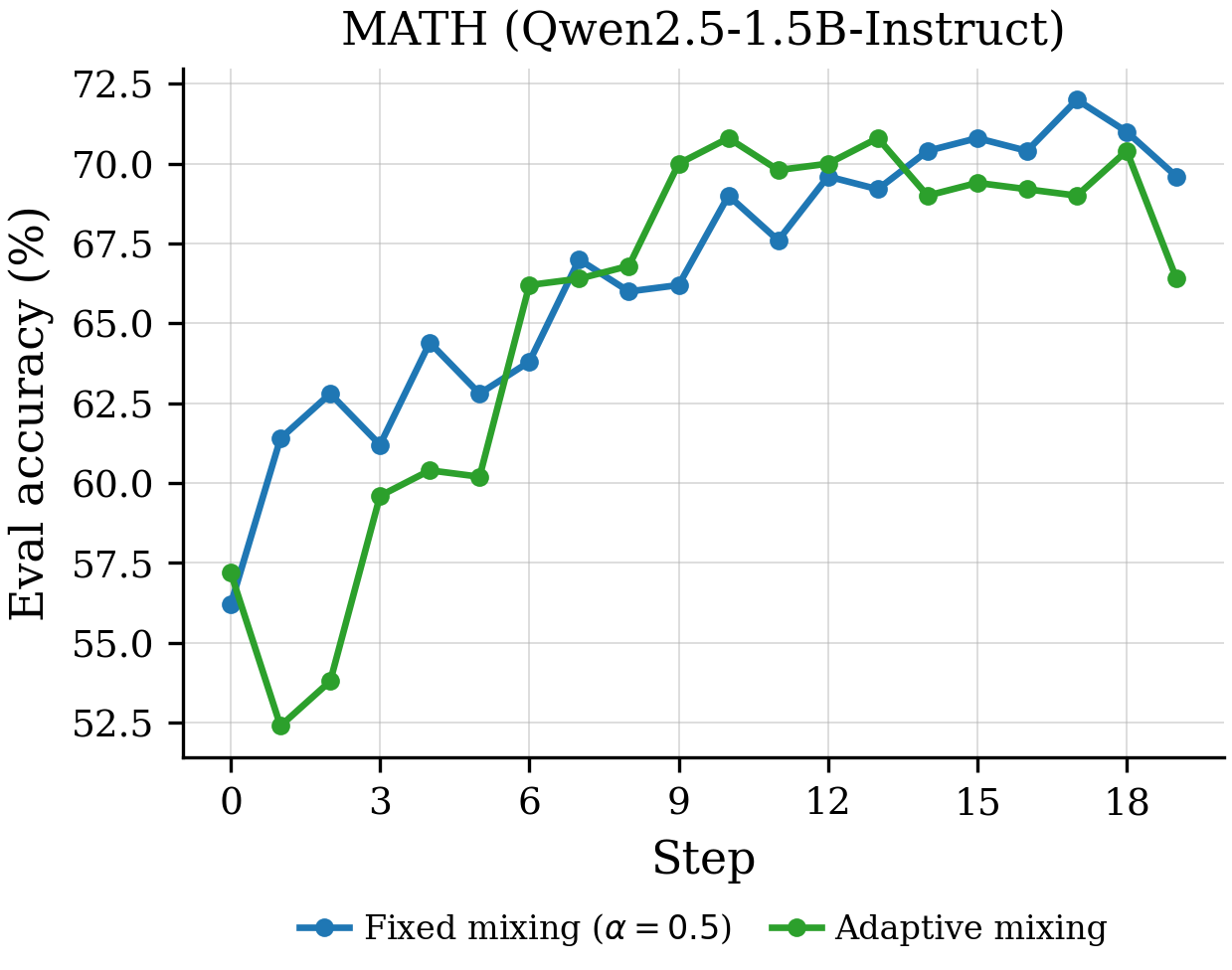}\hfill
\includegraphics[width=0.32\linewidth]{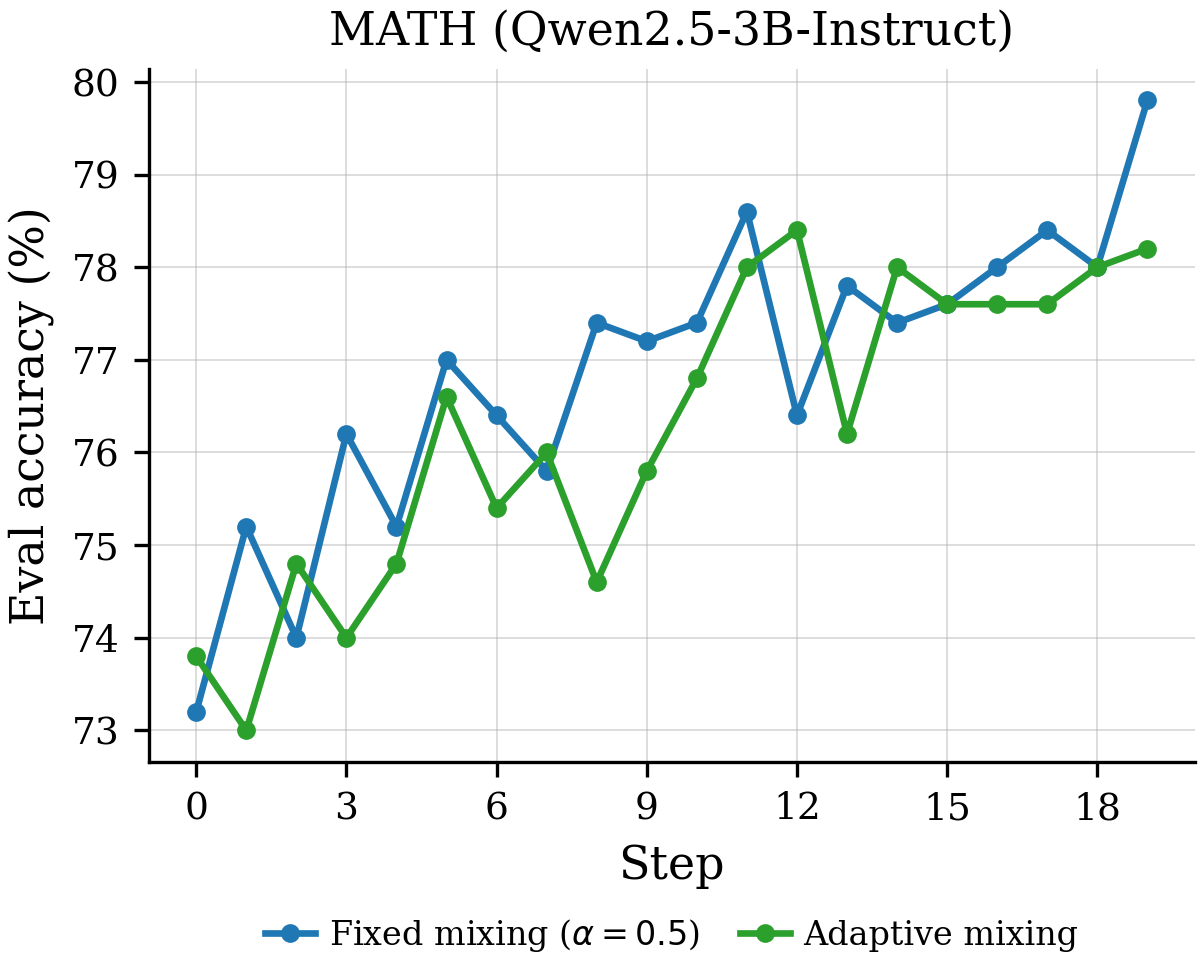}\hfill
\includegraphics[width=0.32\linewidth]{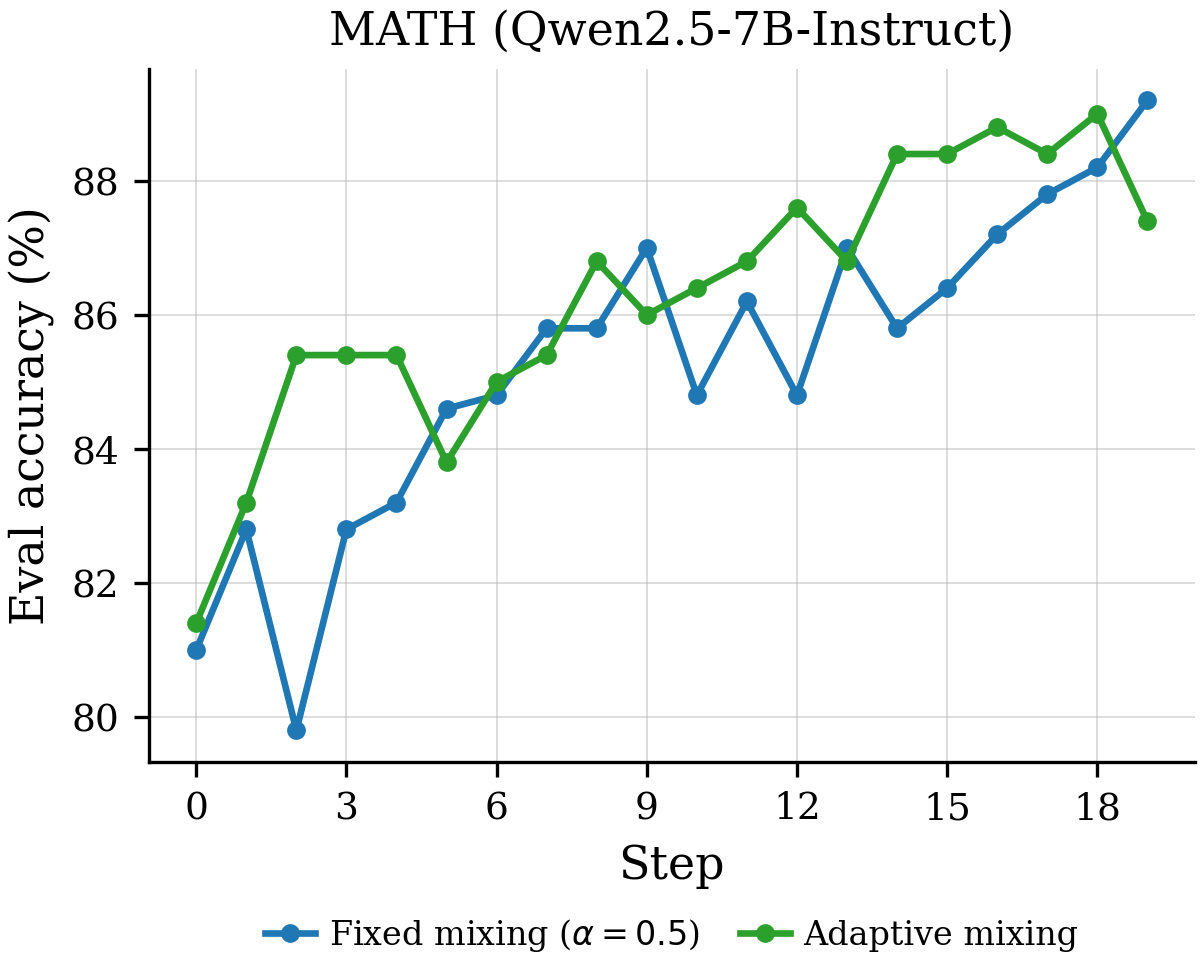}
\caption{MATH: Fixed-mixing LAPO ($\alpha{=}0.5$) vs.\ Adaptive-mixing LAPO.}
\label{fig:emp-math-adaptive}
\end{figure}

\begin{figure}[h]
\centering
\includegraphics[width=0.45\linewidth]{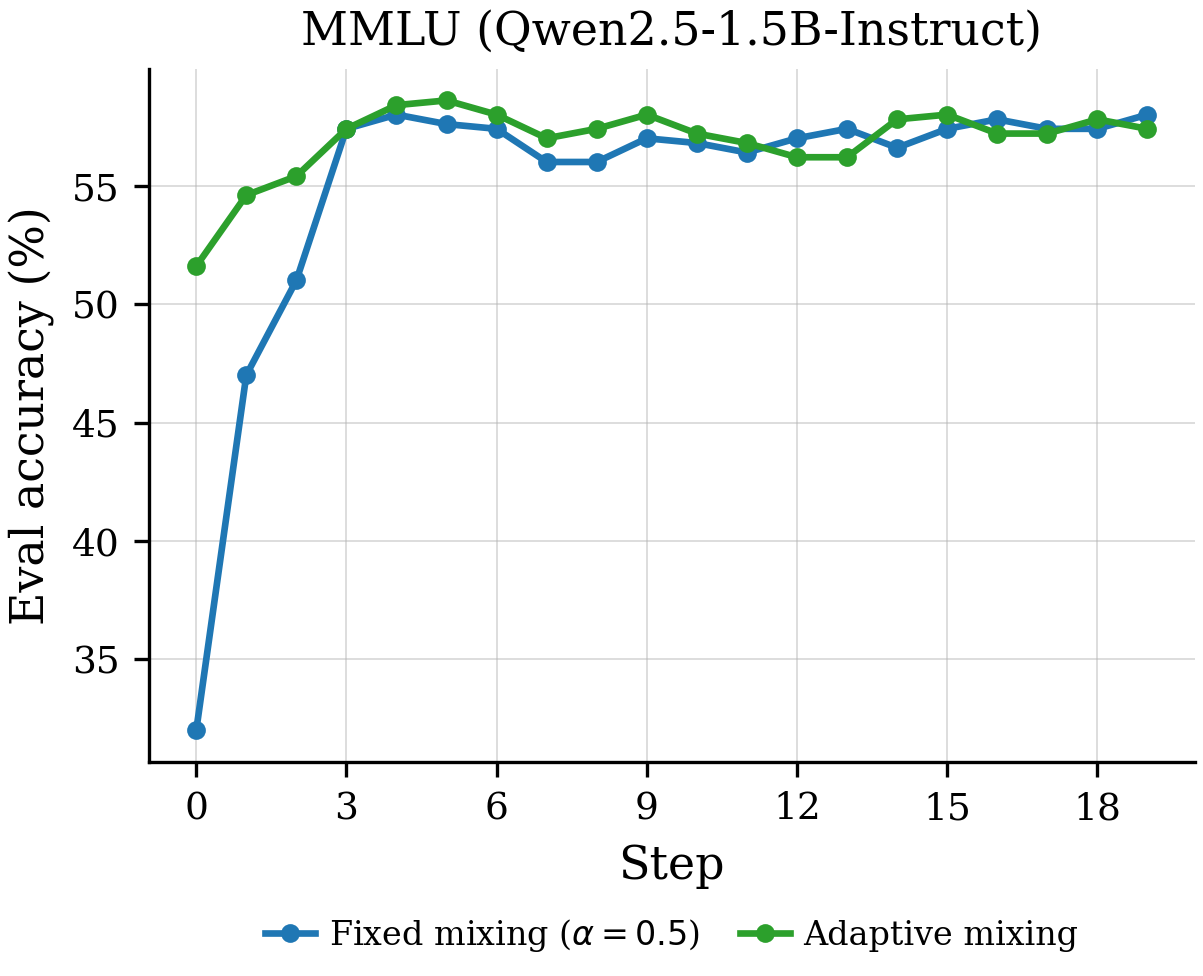}\hfill
\includegraphics[width=0.45\linewidth]{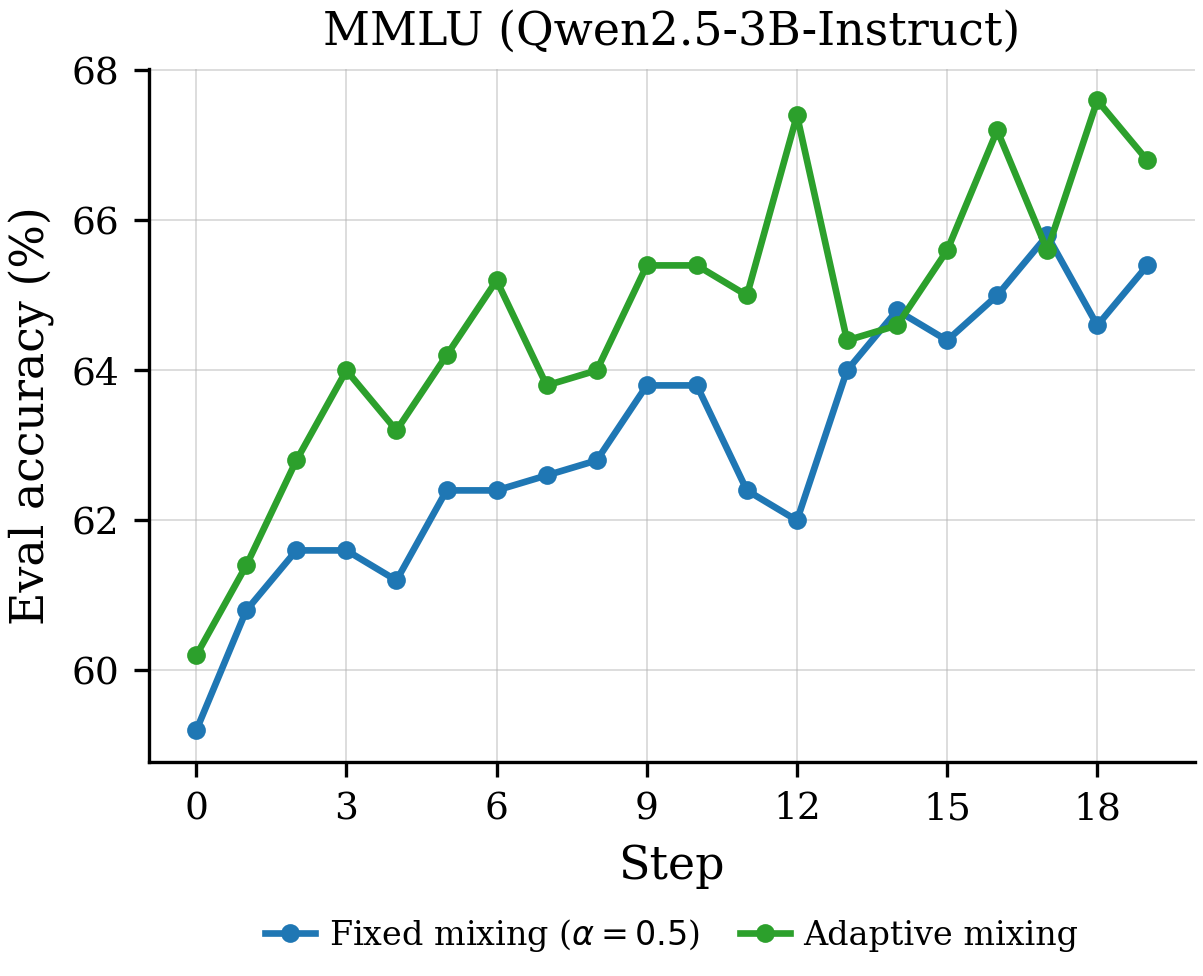}
\caption{MMLU: Fixed-mixing LAPO ($\alpha{=}0.5$) vs.\ Adaptive-mixing LAPO.}
\label{fig:emp-mmlu-adaptive}
\end{figure}

\begin{figure}[h]
\centering
\includegraphics[width=0.32\linewidth]{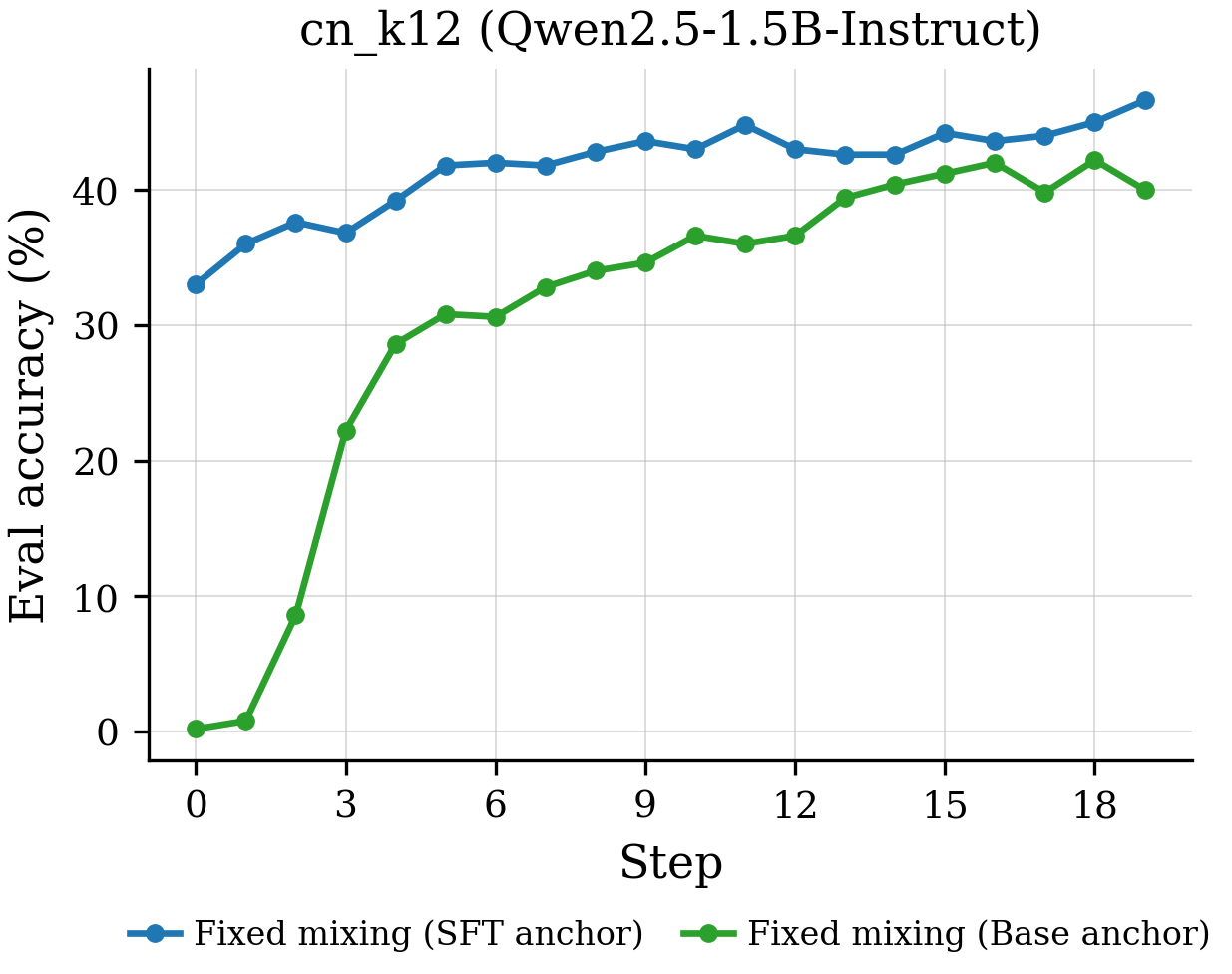}\hfill
\includegraphics[width=0.32\linewidth]{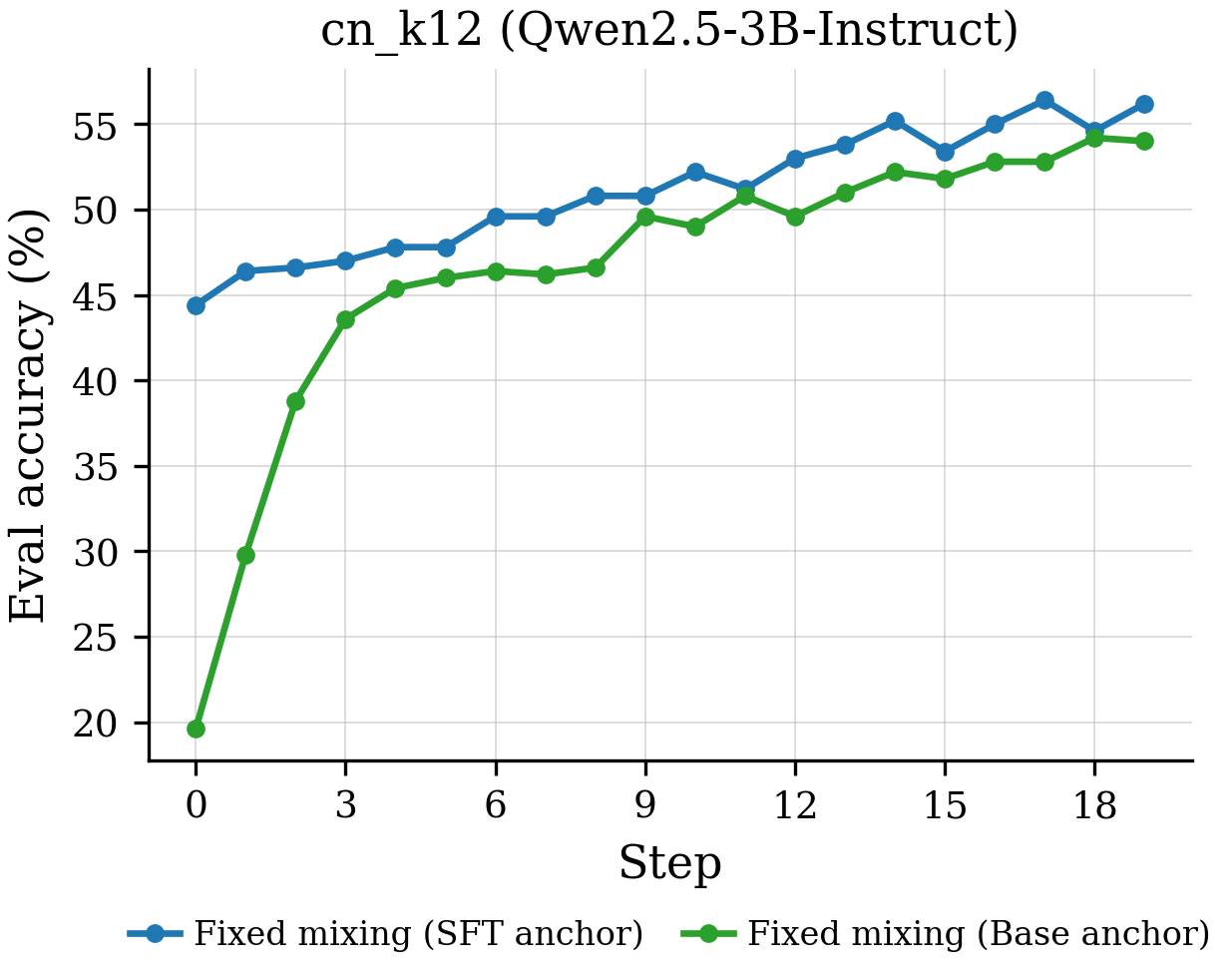}\hfill
\includegraphics[width=0.32\linewidth]{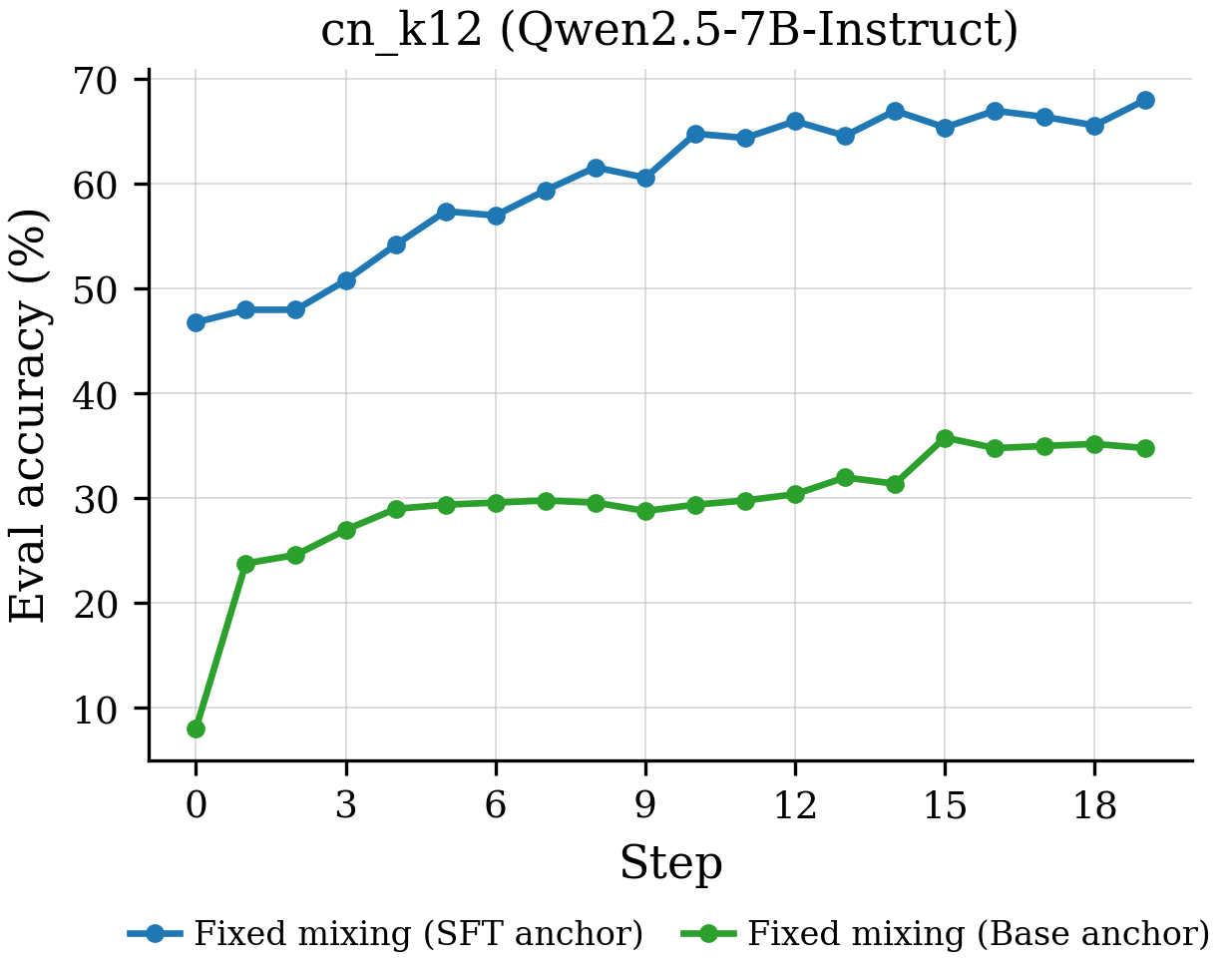}
\caption{cn\_k12 anchor ablation: Fixed-mixing LAPO with SFT anchor vs.\ with base anchor.}
\label{fig:emp-cnk12-base}
\end{figure}


\begin{table}[h]
\centering
\caption{Format rate and solve rate on cn\_k12 (500-problem held-out set) by varying the weight $\alpha$ in logit averaging ($z^{\mathrm{mix}}_{\theta,\alpha}\!=\!(1{-}\alpha)z_{\theta_{\mathrm{base}}}+\alpha z_{\mathrm{ref}}$, Eq.~\eqref{eq:logit-mix}) and probability averaging ($\pi^{\mathrm{prob}}_{\theta,\alpha}\!=\!(1{-}\alpha)\pi_{\theta_{\mathrm{base}}}+\alpha\pi_{\mathrm{ref}}$); $\alpha{=}0$ gives the pure base policy and $\alpha{=}1$ gives pure SFT. Bold marks the best solve rate within each scheme.}
\label{tab:logit-vs-prob}
\begin{tabular}{lcccc}
\toprule
 & \multicolumn{2}{c}{Logit mixing} & \multicolumn{2}{c}{Prob mixing} \\
\cmidrule(lr){2-3}\cmidrule(lr){4-5}
$\alpha$ & Fmt & Acc & Fmt & Acc \\
\midrule
$0.0$ \;
($\pi_{\theta_{\mathrm{base}}}$)  & $0.0\%$  & $0.0\%$  & $0.0\%$  & $0.0\%$ \\
$0.1$              & $0.4\%$  & $0.2\%$  & $0.0\%$  & $0.0\%$ \\
$0.2$              & $5.8\%$  & $3.8\%$  & $0.4\%$  & $0.4\%$ \\
$0.3$              & $32.0\%$ & $17.6\%$ & $1.0\%$  & $0.6\%$ \\
$0.4$              & $80.2\%$ & $50.4\%$ & $5.2\%$  & $3.0\%$ \\
$0.5$              & $89.2\%$ & $\mathbf{55.6\%}$ & $75.0\%$ & $47.2\%$ \\
$0.6$              & $81.6\%$ & $47.6\%$ & $88.4\%$ & $\mathbf{50.6\%}$ \\
$0.7$              & $87.8\%$ & $45.4\%$ & $91.0\%$ & $46.2\%$ \\
$0.8$              & $88.2\%$ & $45.4\%$ & $90.2\%$ & $44.0\%$ \\
$0.9$              & $92.2\%$ & $43.4\%$ & $90.2\%$ & $40.4\%$ \\
$1.0$ \;
($\pi_{\mathrm{sft}}$) & $90.8\%$ & $39.8\%$ & $90.8\%$ & $39.8\%$ \\
\bottomrule
\end{tabular}
\begin{minipage}{\textwidth}
\footnotesize
\bigskip
Here, $\pi_{\theta_{\mathrm{base}}}$ refers to the policy \emph{before} training, i.e., the base Qwen2.5-1.5B-Instruct model itself (the starting $\theta$, not the final RL-trained $\theta$). Only the mixing weight $\alpha$ and the choice of mixing scheme (logit vs. probability) are varied across the rows of Table~\ref{tab:logit-vs-prob}. This isolates the effect of logit vs. probability averaging from the effect of RL training.
\end{minipage}
\end{table}

\begin{figure}[h]
\centering
\includegraphics[width=0.65\linewidth]{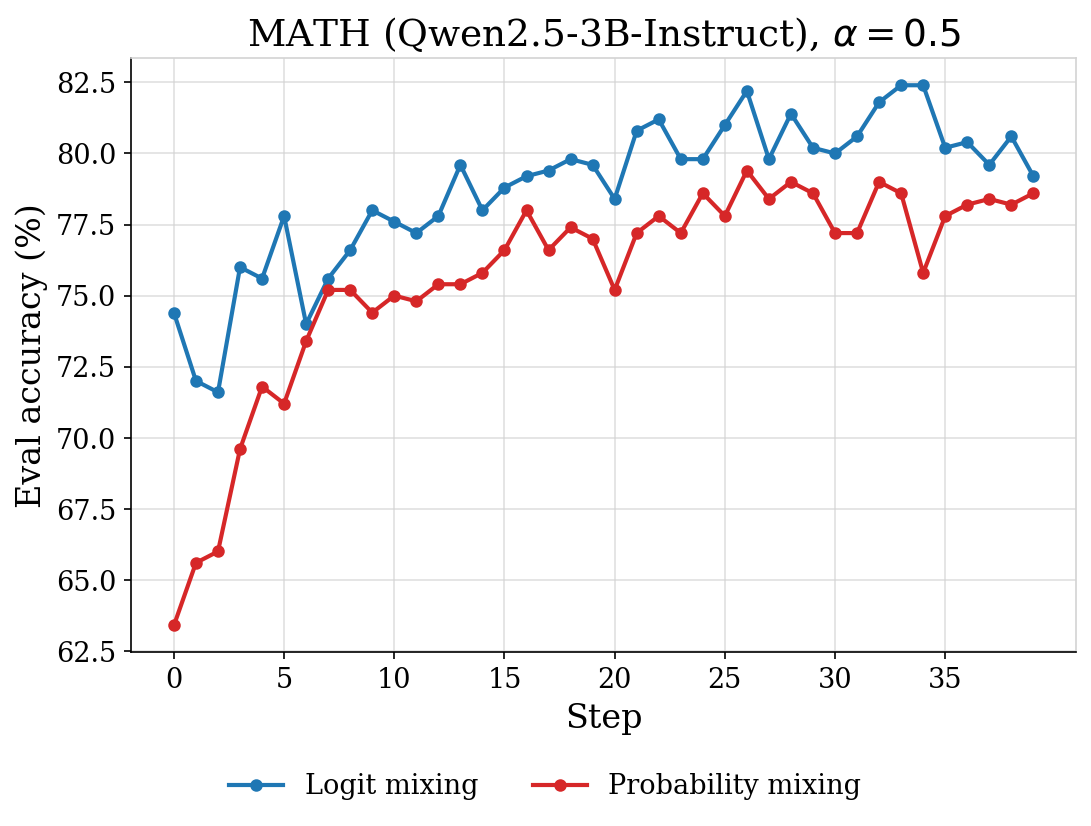}
\caption{Fixed-mixing LAPO vs. probability mixing on MATH for Qwen2.5-3B with $\alpha{=}0.5$.}
\label{fig:logit-vs-prob-training}
\end{figure}

We highlight three important observations in Table~\ref{tab:logit-vs-prob}. \emph{(i) Logit averaging sees improvement with much smaller weight.} In particular, format rate climbs to $32\%$ at $\alpha{=}0.3$ and to $80\%$ at $\alpha{=}0.4$ for logit averaging, while staying below $5\%$ until $\alpha{=}0.4$ and then reaching $75\%$ at $\alpha{=}0.5$ for probability averaging. This contrast lies in the difference between a geometric mean and an arithmetic mean.
\emph{(ii) Logit averaging peaks higher.} The best solve rate for logit averaging ($55.6\%$ at $\alpha{=}0.5$) beats the best solve rate ($50.6\%$ at $\alpha{=}0.6$) for probability averaging, and both exceed pure SFT ($39.8\%$). \emph{(iii) The respective peak sits on the opposite side of $\alpha{=}0.5$.} Logit averaging peaks at $\alpha{=}0.5$ and deteriorates for larger $\alpha$, while probability averaging peaks at $\alpha{=}0.6$. This is because a geometric mean extracts useful signals from $\pi_{\theta_{\mathrm{base}}}$ even when the reference is given equal weight, while an arithmetic mean 
obtains the highest accuracy only when the reference starts to dominate. The same gap persists when the two mixing schemes are used 
inside 
Algorithm~\ref{alg:fixed}.

In view of Figure \ref{fig:logit-vs-prob-training}, logit mixing outperforms probability mixing at essentially every training step: it starts about $11$ points higher
and remains $2$--$5$ points ahead for the rest of training.
This dynamics reinforces the observations made in Table~\ref{tab:logit-vs-prob}, and the gain we observe is specifically tied to the geometric form of logit averaging, instead of the mere interpolations between two policies.


\begin{figure}[h]
\centering
\includegraphics[width=0.49\linewidth]{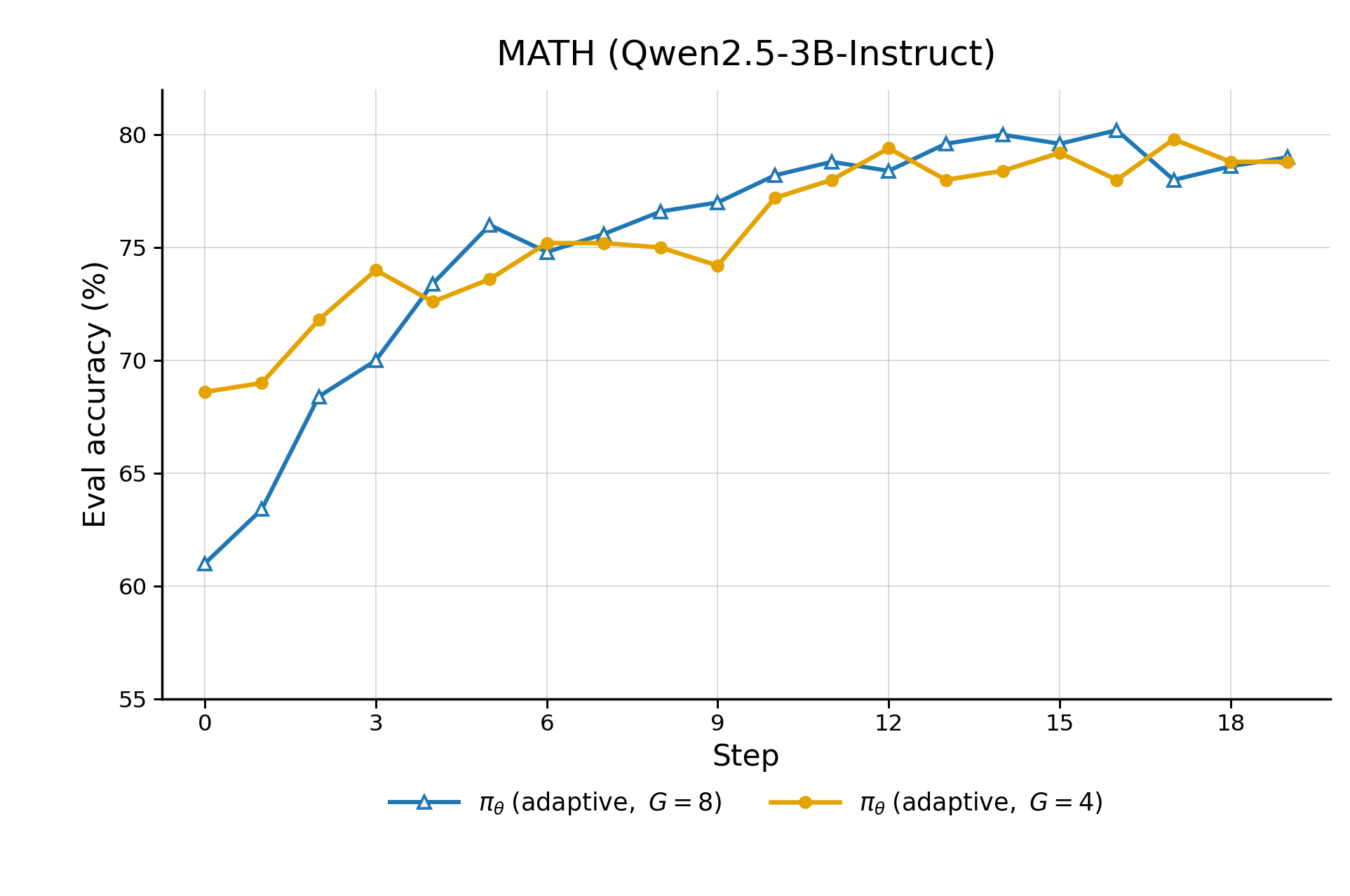}\hfill
\includegraphics[width=0.49\linewidth]{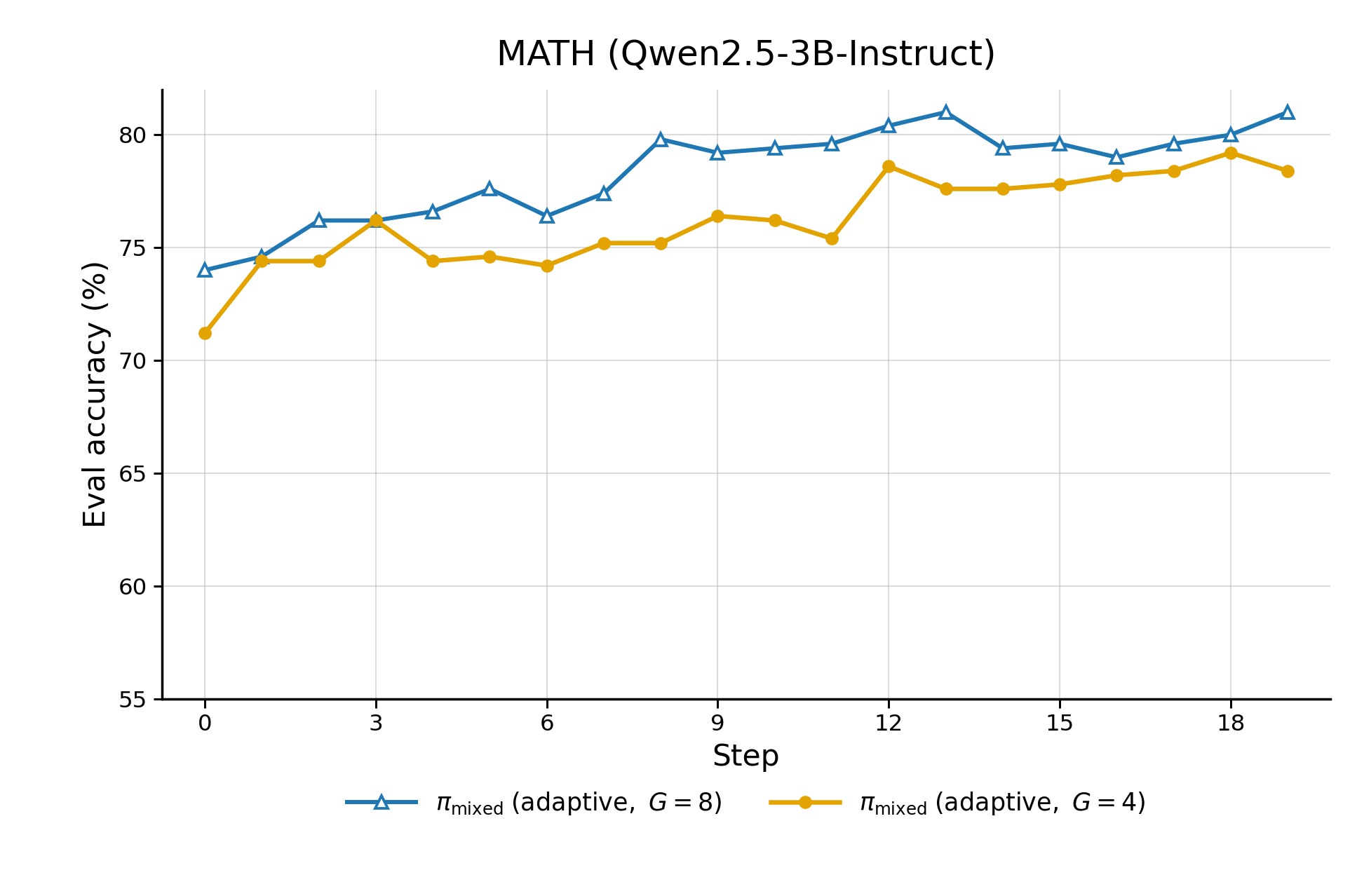}
\caption{Effect of rollout group size on MATH (Qwen2.5-3B-Instruct): evaluation accuracy of the trainable policy $\pi_{\theta}$ (left) and the mixed policy $\pi_{\mathrm{mixed}}$ (right) from Algorithm \ref{alg:adaptive} for $G=4$ and $G=8$.}
\label{fig:math-group-size}
\end{figure}

\end{document}